\documentclass{article} 
\usepackage{iclr2027_conference,times}

\usepackage{amsmath,amsfonts,bm}

\def\eqref#1{equation~\ref{#1}}

\def\1{\bm{1}}

\DeclareMathAlphabet{\mathsfit}{\encodingdefault}{\sfdefault}{m}{sl}
\SetMathAlphabet{\mathsfit}{bold}{\encodingdefault}{\sfdefault}{bx}{n}

\usepackage{hyperref}
\usepackage{url}

\usepackage{graphicx}
\usepackage{comment}
\usepackage{hyperref}
\usepackage{algorithm}
\usepackage{algpseudocode}
\usepackage{booktabs}
\usepackage[switch]{lineno}
\usepackage{enumitem}
\usepackage{xurl}
\usepackage{subcaption}
\usepackage{wrapfig}
\usepackage{amssymb}

\usepackage{amsmath}

\definecolor{darkgreen}{RGB}{0,128,0}

\newcommand{\system}{\texttt{FlowMaker}}
\newcommand{\FMNoReuse}{\texttt{FlowMakerNoReuse}}

\newcommand{\wfvar}{workflow-variant}

\title{{\system}: From Tasks to Cost-Efficient Generative AI Workflow Deployments}

\author{Dimitrios Liakopoulos, Prasoon Sinha, Neeraja J. Yadwadkar\\
The University of Texas at Austin\\
}

\iclrfinalcopy 
\begin{document}

\maketitle

\begin{abstract}
Generative AI workflows execute user tasks by composing sequences of LLM calls, where each call performs a sub-task (e.g., summarization or translation), forming a workflow. Executing such workflows requires decisions over workflow structure, model assignments, and deployment configurations, which jointly determine latency, cost, and accuracy under dynamic workloads and resource constraints. Existing systems optimize these decisions in isolation or fix them early, leading to suboptimal cost and goodput. We make two key observations. First, efficient serving requires jointly exploring all combinations of workflows, model choices, and deployment configurations under dynamic cluster state. Second, reuse of deployed LLMs is critical for reducing cost, but must be performed on-the-fly, as reuse depends on current load and resource availability. Together, these make selecting a {\wfvar}—a combination of workflow, models, and deployment configuration—a time-critical problem. We present FlowMaker, which exposes a task-level interface and selects {\wfvar}s on behalf of users. To realize this, we design a constraint-guided load-aware pruning algorithm for fast, on-the-fly decisions. FlowMaker improves goodput by up to 10.14$\times$ and reduces cost by up to 2.77$\times$ compared to state-of-the-art baselines, including Murakkab.
\end{abstract}

\vspace{-4mm}
\section{Introduction}
\label{sec:intro}
\vspace{-2mm}

Generative AI applications increasingly execute complex user tasks by composing multiple LLM calls into workflows~\citep{BAIR_CompoundAI, ScaffoldedLLMs, Murakkab, AWS_Quick_Flows, Nalar, AFlow, LangChain, LangGraph, CrewAI, AutoGen, Google_Vertex_Agent_Engine, Azure_AI_foundry}.
Consider answering questions over enterprise data when relevant documents must be summarized and the result translated.
\begin{wrapfigure}{r}{0.5\linewidth}
\vspace{-2mm}
    \centering
    \includegraphics[width=\linewidth]{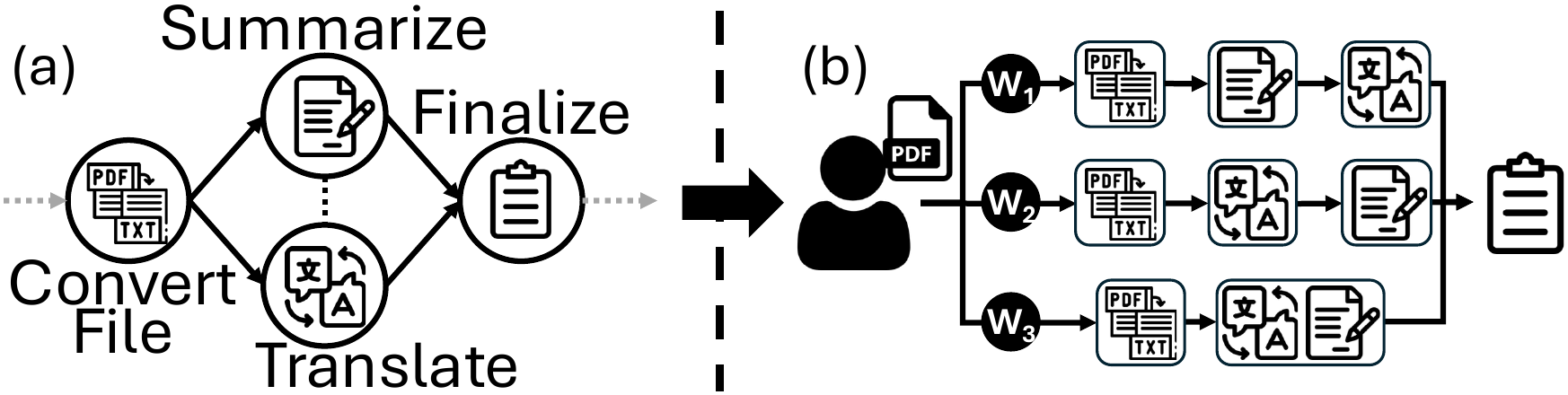}
    \vspace{-7mm}
    \caption{\small{
    (a) Task graph for document conversion, summarization, and translation, showing the sub-tasks and their dependencies.
    (b) The three candidate workflows considered in our example. $W_1$ converts the user-provided PDF to text, summarizes it, and then translates the summary; $W_2$ and $W_3$ use alternative organizations of the sub-tasks.
    Assigning models and deployment configurations to these workflows yields over 24K variants, with significant variation in latency and cost. 
    }}
    \label{fig:workflow_example}
    \vspace{-2mm}
\end{wrapfigure}
This task can be executed by (a) summarizing the documents and then translating the summary, or (b) translating the documents before summarization, or (c) performing summarization and translation jointly.
After choosing a workflow, the system must assign a model to each LLM sub-task (e.g., Llama-3.1-8B or Qwen-2.5-7B) and select a deployment configuration for each model (e.g., tensor- and pipeline-parallelism settings).
We call each combination of workflow, model assignments, and deployment configurations a \emph{\wfvar}.
For the task in Figure~\ref{fig:workflow_example}, assigning 11 models and 10 deployment configurations to the LLM sub-tasks in three candidate workflows yields over $24K$ {\wfvar}s.
Across these variants, latency varies by $189.3\times$ ($0.6$–$119$s), cost by $8725\times$ ($2\times10^{-5}$–$0.17$ \text{\textcent}), and accuracy by $3.3\times$ ($14.4$–$48.0\%$). 
Thus, the selected {\wfvar} directly affects the quality, performance, and cost of serving the task.

Being multi-dimensional, these decisions are already complex to make. 
This complexity is further exacerbated by dynamic conditions: Task arrival rates vary over time~\citep{ALAP, AQUATOPE, Cypress}, workload mixes evolve~\citep{BurstGPT, Splitwise, DynamoLLM, ModServe}, and resource availability changes~\citep{resource_availability_1, ResourceCentral}.
However, existing workflow frameworks typically fix these decisions statically~\citep{LangChain, LangGraph, CrewAI, AutoGen, MetaGPT}, while managed platforms leave them to manual tuning~\citep{AWS_Quick_Flows, Google_Vertex_Agent_Engine, Azure_AI_foundry}.
Across our experiments, such fixed choices incur up to $4.00\times$ the serving cost ($2.93\times$ on average) and achieve up to $12.84\times$ lower goodput than a system that jointly adapts these decisions (\S~\ref{sec:eval_e2e}).

In this paper, we study how to serve generative AI tasks under dynamic workloads and resource constraints.
Our goal is to increase goodput, the rate of requests completed while satisfying their Service-Level Objectives (SLOs), while reducing serving cost.
Existing systems optimize only subsets of the decisions that determine a {\wfvar}~\citep{LangChain, LangGraph, CrewAI, AutoGen, Google_Vertex_Agent_Engine, Azure_AI_foundry, AFlow, LLMSelector, RouteLLM, MaverIQ, Vidur, AlpaServe}.
AFlow~\citep{AFlow} focuses on task decomposition; LLMSelector~\citep{LLMSelector} and RouteLLM~\citep{RouteLLM} focus on model selection; and MaverIQ~\citep{MaverIQ}, Vidur~\citep{Vidur}, and AlpaServe~\citep{AlpaServe} support deployment selection.
Murakkab~\citep{Murakkab}, the closest prior system, jointly considers model and deployment choices but fixes the user-provided workflow.
Although effective within their respective dimensions, these approaches do not jointly revisit workflows, model assignments, and deployment configurations as workload and cluster conditions change.
Consequently, they fail to capture opportunities that arise only when these decisions are made jointly: particularly reuse of LLMs across tasks and adaptation to the current cluster state (e.g., already deployed models, load, and resource availability). 
We find that this leads to suboptimal performance and resource utilization (\S~\ref{sec:full-search-space}--\ref{sec:reuse}). 
For example, when serving seven tasks, we observed that an oracle that searches across all three dimensions reduces serving cost by $1.85\times$ and increases goodput by $1.69\times$ relative to Murakkab (\S~\ref{sec:full-search-space}).

We argue that serving systems need to make these decisions on behalf of users, rather than relying on statically fixed choices or requiring manual configuration. We propose a \emph{task-level interface}: users specify only the task along with accuracy and performance SLOs, and the system selects a \emph{\wfvar}—a combination of workflow, model choices, and deployment configurations. To support this interface, the serving system needs to \emph{explore the full design space of {\wfvar}s and select one per task} to improve serving cost and goodput.

Joint selection is especially important as different tasks can reuse the same deployed LLM.
When a suitable model is already deployed and lightly loaded, reusing it can avoid the cost of an additional deployment.
The same decision may violate latency SLOs when that model is heavily utilized.
Because deployed models, their load, and available resources change as tasks arrive, reuse decisions cannot be fixed offline.
Instead, the serving system must select a {\wfvar} and make \emph{on-the-fly reuse} decisions to determine which deployed models can be reused based on the current cluster state.
This turns {\wfvar} selection into an online constrained-search problem.

However, existing off-the-shelf algorithms are not well-suited for this setting. 
For the task in Figure~\ref{fig:workflow_example}, exhaustively evaluating the variants induced by three workflows, 11 models, and 10 deployment configurations per model consumes approximately $40$ CPU-minutes.
On a single CPU, this is approximately $4.7\times$ the task's most relaxed latency SLO of $\sim510$s.
Recent profilers~\citep{MaverIQ, Vidur, MAD_Max, NeuSight} reduce the cost of profiling individual LLM configurations, but do not eliminate the cost of evaluating workflows composed of multiple models.
General-purpose methods such as Bayesian optimization~\citep{BO_org, BO_textbook, AQUATOPE, Composer} and evolutionary search~\citep{evolutionary_textbook, evolutionary_vis} also rely on repeated candidate evaluations.
In our setting, this evaluation cost must be paid whenever changing load or reuse opportunities alter the feasible choices.
Thus, meeting the online decision budget requires exploiting the structure of this problem to eliminate candidates before evaluating complete {\wfvar}s.

To this end, we build {\system}, an inference serving system for generative AI workflows that meets user SLOs while minimizing serving cost. {\system} exposes the task-level interface: users specify only the task along with any latency and accuracy SLOs.
{\system} generates candidate workflows for each task and lets users select the suitable ones.
Involving users in selecting among the generated candidate workflows is intentional for two reasons: (1) task decomposition is often ambiguous and can yield multiple valid interpretations~\citep{planning_dark}, and (2) current LLMs are unreliable as autonomous planners~\citep{planning_ability_1, planning_ability_2, planning_ability_3}.
{\system} then selects a {\wfvar} for each task by jointly reasoning over workflows, model choices, deployment configurations, and on-the-fly reuse under dynamic cluster state.
To make this search fast, we design a novel load-aware pruning algorithm that progressively narrows the candidate space using accuracy feasibility, operational cost, and performance under the current load.
It then selects a low-cost {\wfvar} estimated to satisfy the user's latency and accuracy SLOs, when one exists, while considering opportunities to reuse already deployed models.

We make the following contributions:
\begin{itemize}[noitemsep,topsep=0pt,leftmargin=*]
\item We formulate serving generative AI workflows as a joint selection problem over workflows, model assignments, and deployment configurations. We show that considering all three dimensions and making reuse decisions from the current cluster state materially improve goodput and serving cost.

\item We design a load-aware {\wfvar} selection algorithm that searches across this joint space without exhaustively evaluating variants. In our evaluation, it reduces search time by $3.7\times$ relative to Aquatope's~\citep{AQUATOPE} Bayesian optimization while accounting for reuse across tasks.

\item We implement {\system} atop TensorRT-LLM. Across loads and arrival patterns, {\system} increases goodput by $1.45$--$10.14\times$ ($4.73\times$ avg) and reduces serving cost by $1.56$--$2.77\times$ ($1.95\times$ avg) over Murakkab~\citep{Murakkab}. Joint search alone improves average goodput and cost by $4.34\times$ and $1.48\times$, respectively, while on-the-fly reuse further reduces cost by $1.31\times$.
\end{itemize}
\vspace{-5mm}
\section{Problem Setting and Motivation}
\label{sec:section2}
\vspace{-4mm}

We first define the decisions involved in deploying a generative AI task. We then show why effective serving requires (1) jointly selecting workflows, models, and deployment configurations and (2) deciding whether to reuse deployed LLM instances based on the current cluster state.

\vspace{-1mm}
\noindent\textbf{Task, request, and workflow.}
A \emph{task} is a high-level user objective accomplished through one or more stages, including LLM calls. A \emph{request} is an invocation of a deployed task with a concrete input. A \emph{workflow} is an executable decomposition of a task into stages, their dependencies, and execution order. A stage may be a non-LLM operation (e.g., PDF-to-text conversion) or an LLM call (e.g., summarize or translate). Figure~\ref{fig:workflow_example}(a) shows the stages and relationships for our example task;  Figure~\ref{fig:workflow_example}(b) shows three candidate workflows with different decompositions or execution orders.

\vspace{-1mm}
\noindent\textbf{Workflow variant.}
A workflow can be realized using different models and deployment configurations. A \emph{\wfvar} specifies a workflow and, for each LLM stage, the model used to execute it (e.g., Llama-3.1-8B or Qwen-2.5-7B) and the model's deployment configuration (e.g., its tensor- and pipeline-parallelism degrees). Variants of the same workflow can therefore differ in accuracy, latency, memory consumption, and serving cost. 
For the example in Figure~\ref{fig:workflow_example}, evaluating 3 workflows, 11 models, and 10 deployment configurations generates $\sim24K$ variants across which latency varies by $189.3\times$ ($0.6$–$119$s), cost by $8725\times$ ($2\times10^{-5}$–$0.17$ \text{\textcent}), and accuracy by $3.3\times$ ($14.4$–$48.0\%$). 

\vspace{-1mm}
\noindent\textbf{Deployment and reuse.}
A selected {\wfvar} must be mapped to model deployments. For each LLM stage, the system can deploy a new model instance or reuse a compatible one already deployed for another task. We define \emph{reuse} as sharing a deployed LLM instance across stages of different tasks. Reuse can reduce serving cost, but adds load to the shared instance and can increase request latency. Thus, reuse is a cluster-state-dependent decision for realizing a {\wfvar}.

\vspace{-1mm}
\noindent\textbf{Problem formulation.}
Given a task, its user-approved candidate workflows, latency and accuracy SLOs, and the current cluster state, the serving system must make two decisions: (1) select a {\wfvar}, and (2) realize that variant by mapping each LLM stage to either a new or a compatible existing model instance. This decision is made when the task is deployed; subsequent requests use the selected deployment. The system seeks a low-cost selection predicted to satisfy both SLOs. We measure its cluster-level effectiveness using \emph{goodput}: the rate of requests that complete within their latency SLO while using a {\wfvar} that satisfies the task's accuracy SLO.

\vspace{-3mm}
\subsection{Why Select Workflows, Models, and Deployment Configurations Jointly?}
\label{sec:full-search-space}
\vspace{-3mm}

\noindent\textbf{Interacting decision dimensions.}
A task may admit multiple user-approved workflows; each LLM stage may use one of several compatible models; and each model can run under multiple tensor- and pipeline-parallelism configurations. These choices cannot be made independently: the workflow determines the stages and dependencies, model assignments determine attainable proxy accuracy and resource requirements, and deployment configurations determine latency, capacity, and cost.

In Figure~\ref{fig:workflow_example}, $W_1$ and $W_2$ each contain two LLM stages, whereas $W_3$ performs summarization and translation through one joint LLM stage; PDF conversion is a fixed non-LLM stage. With 11 models and 10 deployment configurations per LLM stage, this yields
$2(11\cdot10)^2+(11\cdot10)=24{,}310$
workflow variants. Fixing any dimension in advance can therefore exclude low-cost variants satisfying the end-to-end SLOs. Across the evaluated models and configurations, latency, memory, cost, and proxy accuracy expose different tradeoffs; Appendix~\ref{app:model_characterization} provides the complete characterization.

Prior systems restrict this joint space by fixing at least one dimension. AFlow~\citep{AFlow} varies workflows, LLMSelector~\citep{LLMSelector} and RouteLLM~\citep{RouteLLM} select models, and MaverIQ~\citep{MaverIQ} and related systems~\citep{Vidur, NeuSight, AlpaServe} support deployment selection. Murakkab~\citep{Murakkab}, the closest prior system, jointly selects models and deployment configurations but assumes a user-provided workflow.

\begin{wrapfigure}{r}{0.4\linewidth}
\centering
\vspace{-7mm}
\includegraphics[width=\linewidth]{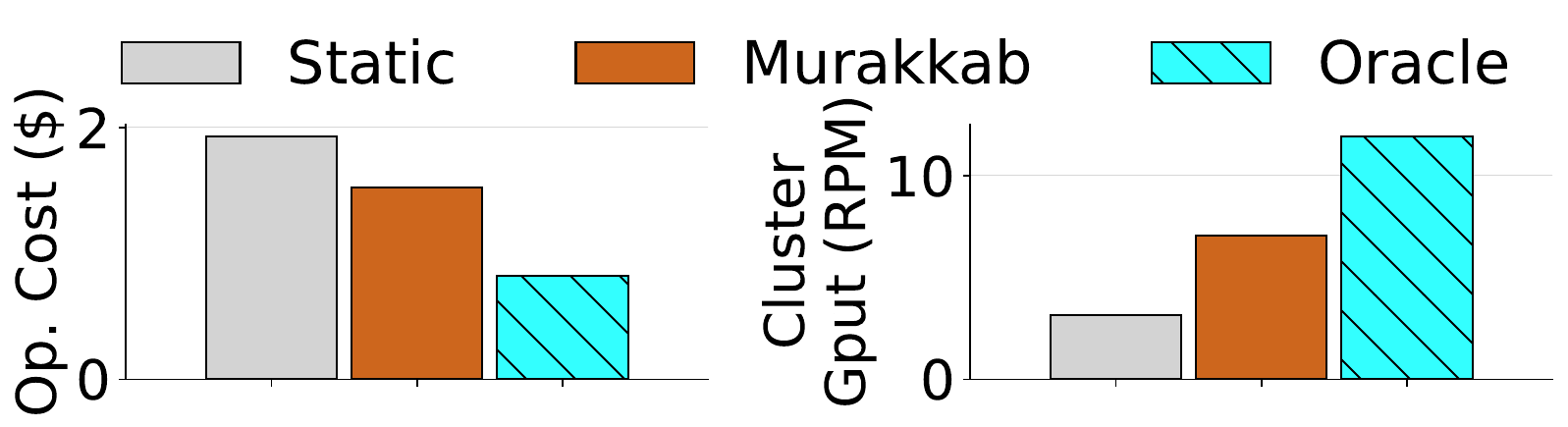}
\vspace{-7mm}
\caption{\small{Serving cost and cluster goodput as the system is given progressively more choices: a fixed workflow variant (Static), model and 
deployment choices within a fixed workflow (Murakkab), and joint workflow, model, and deployment choices from a high-level task (Oracle). Joint optimization reduces cost and improves goodput.
}}
\label{fig:motiv_exploration}
\vspace{-3mm}
\end{wrapfigure}

Figure~\ref{fig:motiv_exploration} quantifies the resulting opportunity. We serve seven tasks under bursty, medium-load request arrivals from the Azure code trace~\citep{Splitwise}. The \emph{Static} policy uses a fixed {\wfvar}; Murakkab receives a default workflow and selects models and deployment configurations; and the \emph{Oracle} exhaustively searches the joint space of candidate workflows, models, and deployment configurations. Murakkab incurs $1.85\times$ the oracle's serving cost, while the oracle achieves $1.69\times$ Murakkab's goodput by also considering alternative workflows.

\vspace{-1mm}
\noindent\textbf{LLM routing alone is insufficient. }
Routing systems~\citep{MasRouter, Aragog, RouteLLM, LLMRouting} select among already-deployed models, whereas {\system} also determines which workflow, models, and deployment configurations should be deployed. So, LLM routing is complementary; for details, see Appendix~\ref{app:related_works}. We also evaluate against RouteLLM~\citep{RouteLLM} in \S~\ref{sec:evaluation}.

\vspace{-1mm}
\noindent\textbf{Requirement 1.} 
The serving system should jointly search workflows, models, and deployment configurations without fixing any dimension in advance, while keeping search efficient. 
Exhaustively evaluating 24,310 {\wfvar}s takes about 40 CPU-minutes, nearly $4.7\times$ the task's most relaxed $\sim510$s latency SLO, making direct enumeration impractical at deployment time.

\vspace{-3mm}
\subsection{Why Reuse Must Account for Current Load?}
\label{sec:reuse}
\vspace{-3mm}

Selecting across workflows and models can expose opportunities for stages from different tasks to share a deployed LLM instance. Such reuse avoids a separate deployment for each task, but its feasibility depends on the load already placed on the shared instance. A deployment that can accommodate another task under low load may violate latency SLOs when its existing request streams approach its SLO capacity---the maximum request rate it can sustain while meeting its latency target. Because request rates and resource availability change over time~\citep{ALAP, AQUATOPE, Cypress, BurstGPT, Splitwise, DynamoLLM, ModServe, resource_availability_1, ResourceCentral}, reuse decisions based on a fixed load estimate can either provision unnecessary capacity or overload shared instances.

\begin{wrapfigure}{r}{0.42\linewidth}
\centering
\vspace{-7mm}
\includegraphics[width=\linewidth]{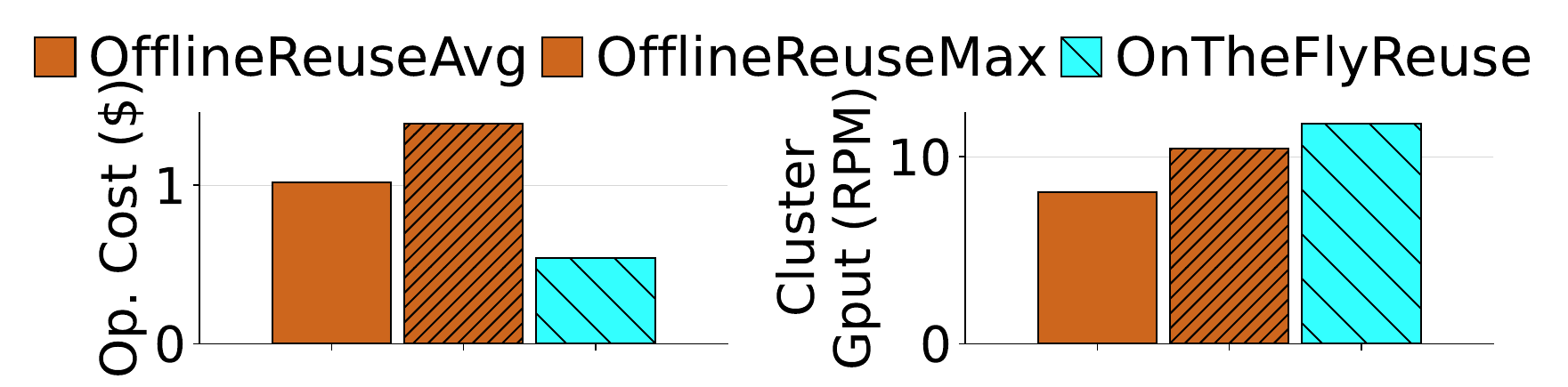}
\vspace{-7mm}
\caption{\small{Serving cost and cluster goodput for reuse policies provisioned for average, maximum, or current cluster load (OnTheFlyReuse). Adapting reuse decisions to current load reduces cost and improves goodput. }
}
\label{fig:reuse}
\vspace{-5mm}
\end{wrapfigure}
Murakkab~\citep{Murakkab} makes reuse decisions offline. We instantiate its offline planner in two ways: \emph{OfflineReuseAvg} provisions shared instances for the average request load, while \emph{OfflineReuseMax} provisions them for the maximum load. We evaluate both while serving seven tasks whose request streams follow the bursty Azure code trace~\citep{Splitwise}. Figure~\ref{fig:reuse} shows that OfflineReuseMax incurs $1.36\times$ OfflineReuseAvg's cost but achieves $1.29\times$ its goodput. Thus, provisioning for average load reduces cost but creates contention during bursts, while provisioning for peak load preserves more capacity at higher cost.
\emph{OnTheFlyReuse} instead makes reuse decisions as a task is deployed, using the current load and available capacity of existing model instances. 
Compared with Murakkab's offline policies, OnTheFlyReuse improves both cost and goodput by $1.89$–$2.57\times$ and $1.13$–$1.45\times$, respectively.

\vspace{-1mm}
\noindent\textbf{Requirement 2.} 
The serving system should therefore make reuse decisions based on the current load and capacity of deployed model instances. 
Together, efficient joint selection and load-aware reuse motivate {\system}'s design in the following section.

\vspace{-5mm}
\section{{\system}}
\label{sec:section3}
\vspace{-3mm}

\begin{wrapfigure}{r}{0.450\linewidth}
\vspace{-3mm}
\centering
\includegraphics[width=\linewidth]{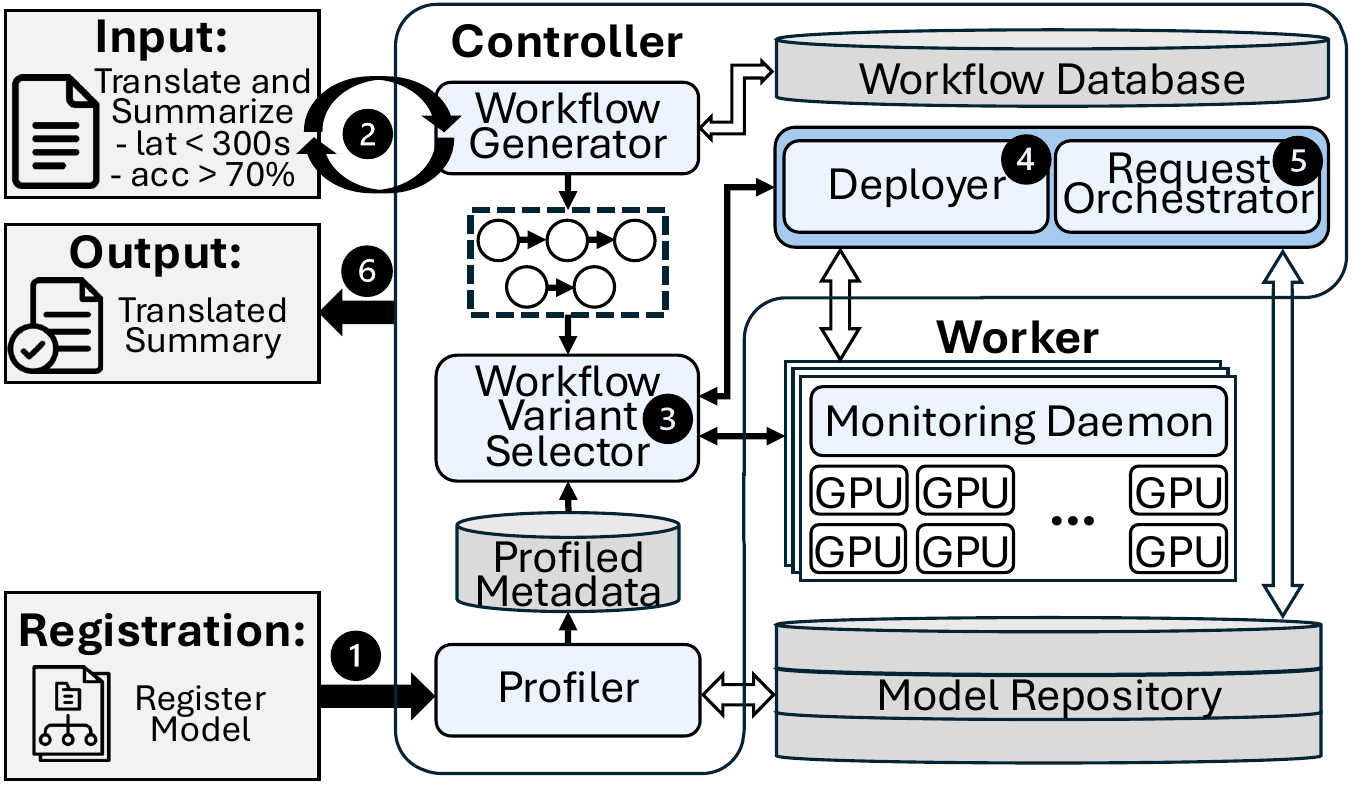}
\vspace{-7mm}
\caption{{\system}. See~\S~\ref{subsec:interface}.}
\label{fig:design}
\vspace{-5mm}
\end{wrapfigure}

The previous section established two requirements: selection must jointly consider workflows, models, and deployment configurations, and reuse decisions must account for the current cluster state. {\system} meets these requirements by separating reusable offline profiling from state-dependent selection at task-deployment time. Figure~\ref{fig:design} shows its architecture.

\vspace{-3mm}
\subsection{Task-Level Interface and System Lifecycle}
\label{subsec:interface}
\vspace{-3mm}

{\system} exposes a task-level interface (see Appendix~\ref{app:implementation_details}-APIs). Through \texttt{deploy$\_$task}, users submit a natural-language description of a task and its accuracy and latency SLOs. {\system} generates candidate workflows, from which the user identifies the semantically acceptable ones. This validation is necessary because task decomposition can admit multiple valid interpretations~\citep{planning_dark}, and current LLMs remain unreliable as autonomous planners~\citep{planning_ability_1, planning_ability_2, planning_ability_3}. {\system} then selects a {\wfvar}, determines which deployed model instances can be reused, and deploys any remaining models. 
Subsequent \texttt{inference$\_$task} requests use this selected deployment. Expert users may add constraints or directly specify workflows, models, or deployment configurations; {\system} respects these choices while still exploiting compatible deployed instances when possible.
{\system}'s lifecycle consists of three steps.

\vspace{-1mm}
\noindent\textbf{Model registration and profiling.}
When a model is registered through \texttt{register$\_$model}, {\system} stores its weights and profiles its supported tensor- and pipeline-parallelism configurations. The Profiler builds on MaverIQ~\citep{MaverIQ} to estimate the latency and memory requirements of each configuration. These profiles are stored and reused across tasks.

\vspace{-1mm}
\noindent\textbf{Task deployment.}
When \texttt{deploy$\_$task} is invoked, the Workflow Generator uses an LLM to decompose the task into candidate workflows and refines them through user interaction (\S~\ref{subsec:interface}). It caches the user-approved workflows and passes them to the Workflow-Variant Selector. The Selector combines these workflows with model profiles, user SLOs, and the current cluster state, including deployed models, their load, and available GPU memory, to select a {\wfvar} and reuse plan. The Deployer then launches only the model instances that cannot be served through reuse.

\vspace{-1mm}
\noindent\textbf{Request execution.}
After deployment, the Orchestrator executes incoming requests using the selected workflow, models, configurations, and reuse mappings. Intermediate outputs are passed according to the workflow dependencies, and the final result is returned to the user. Workflow-variant selection is therefore performed when the task is deployed, rather than for every request.

\vspace{-2mm}
\subsection{Workflow-Variant Selection Problem}
\label{subsec:wf_algo}
\vspace{-2mm}

\vspace{-1mm}
\noindent\textbf{Inputs.}
The selector receives a task $\mathcal{T}$, its accuracy SLO $\alpha$, latency SLO $\lambda$, and expected request arrival rate $\rho$, estimated from historical request traces~\citep{AlpaServe, Murakkab}. It also receives the task's user-approved candidate workflows $\mathcal{W}$, supported models $\mathcal{M}$, and deployment configurations $\mathcal{C}$. Finally, it observes the current cluster state, including the set of deployed model instances $\mathcal{D}=\{n_1,\ldots,n_d\}$, their configurations and serving loads, and available GPU memory.

\vspace{-1mm}
\noindent\textbf{Output.}
For a workflow with $k$ LLM stages, the selector returns a {\wfvar} and its deployment mapping, represented as
$\mathcal{WV}=\langle\mathbf{n},\mathbf{gpu}\rangle$, where
$\mathbf{n}=(n_1,\ldots,n_k)$ specifies each stage's model and deployment configuration, and $\mathbf{gpu}$ specifies its GPU mapping. If a selected node $n_i$ is in $\mathcal{D}$, the stage reuses that model instance; otherwise, {\system} deploys a new instance.

\vspace{-1mm}
\noindent\textbf{Objective and constraints.}
{\system} seeks a low-cost selection that satisfies the task's SLOs without exceeding the capacity of any reused model. For a candidate $\mathcal{WV}$, it computes:

\vspace{-5mm}
\begin{equation}
\mathrm{cost}(\mathcal{WV})=\Big(\sum_{n_i\in\mathbf{n},\,n_i\notin\mathcal{D}}\mathrm{mem}(n_i)\Big)\times\Big(\sum_{n_i\in\mathbf{n}}\mathrm{lat}(n_i)\Big),
\label{eq:cost}
\vspace{-1mm}
\end{equation}
subject to latency, accuracy, and load capacity constraints:

\vspace{-6mm}
\begin{equation*}
\underbrace{\textstyle\sum_{n_i\in\mathbf{n}}\mathrm{lat}(n_i)\le\lambda}_{\text{latency}},\qquad
\underbrace{\min\nolimits_i\,\mathrm{acc}(n_i)\ge\alpha}_{\text{accuracy}},\qquad
\underbrace{\mathrm{cap}(n_i)\,\ge\mathrm{load}(n_i)+\sigma_i\cdot\rho}_{\text{capacity}},
\end{equation*}
\vspace{-1mm}
where $\sigma_i$ counts the subtasks of $\mathcal{WV}$ routed through $n_i$ and $\mathrm{load}(n_i)=0$ for new nodes. The capacity constraint ensures that a deployed model is shared only if the added load keeps it below the rate at which queuing begins.
A candidate must also admit a placement under the GPUs' available memory. We define $\text{cost}(\varnothing)=\infty$.

\vspace{-1mm}
\noindent\textbf{Estimating workflow accuracy.}
To our knowledge, no general method estimates the end-to-end accuracy of an arbitrary generative AI workflow from per-model accuracies before execution. Following prior work~\citep{MaverIQ}, we use the average score across six Open LLM Leaderboard benchmarks~\citep{OpenLLM_leaderboard} as each model's accuracy estimate. Motivated by evidence that per-stage and end-to-end accuracy are correlated~\citep{LLMSelector, Aragog}, we approximate workflow accuracy as the minimum accuracy across its stages. More advanced task- or workflow-specific estimators can be incorporated without changing the remainder of the selector.

\vspace{-2mm}
\subsection{Constraint-Guided Workflow-Variant Selection}
\label{subsec}
\vspace{-2mm}

Directly enumerating all {\wfvar}s is too expensive at task-deployment time. As shown in \S~\ref{sec:full-search-space}, even our translation--summarization example contains 24,310 variants and requires approximately 40 CPU-minutes to evaluate with MaverIQ~\citep{MaverIQ}. Precomputed lookup tables also cannot directly capture changing reuse opportunities and model loads. 
Evolutionary search and Bayesian optimization, often used to manage resources for CPU-based workflows deployed on serverless platforms~\citep{AQUATOPE, Bilal-Serverless}, avoid complete enumeration of the search space. 
However, for generative AI inference serving, they still require repeated evaluations of complete candidate configurations and introduce additional modeling or tuning overhead~\citep{BO_org, BO_textbook, evolutionary_textbook}. 
We further discuss these alternatives, along with graph-based formulations, in Appendix~\ref{sec:appendix_alog_alternatives} and why generative AI workflows differ from traditional serverless workflows in Appendix~\ref{app:related_works}.

{\system} instead exploits when different constraints become evaluable. Accuracy can be checked after choosing models; incremental deployment footprint can be evaluated after constructing a model assignment and accounting for reuse; and latency, capacity, and memory feasibility can be evaluated after selecting deployment configurations and placement. {\system} orders its search according to this dependency structure, applying inexpensive constraints before expanding the next decision dimension.
Algorithm~\ref{algo:wvs} implements this strategy in  three stages. It prioritizes finding a low-cost feasible {\wfvar} quickly over slow yet globally optimal selection.

\begin{algorithm}[t]
\algtext*{EndIf}
\algtext*{EndFor}
\algtext*{EndWhile}
\algtext*{EndProcedure}
\algtext*{EndFunction}
\algnewcommand\algorithmicinput{\textbf{Input:}}
\algnewcommand\Input{\item[\algorithmicinput]}
\algrenewcommand\algorithmicensure{\textbf{Output:}}
\caption{Workflow-variant selection}
\label{algo:wvs}
\small
\begin{algorithmic}[1]
\Input Task $\mathcal{T}$, accuracy SLO $\alpha$, latency SLO $\lambda$, expected load $\rho$, model set $\mathcal{M}$, configuration set $\mathcal{C}$
\Ensure Workflow-variant selection $\mathcal{WV}^{*}$, or $\varnothing$
\State $\mathcal{WV}^{*} \leftarrow \varnothing$
\For{each workflow $[s_1,\ldots,s_k]$ in $\textsc{GetWorkflows}(\mathcal{T})$}
  {\setlength{\fboxsep}{1pt} \Statex \Comment{---------------------------------------------\fbox{\textit{Stage 1: Pruning by accuracy}}-------------------------------------------}}
  \State $\mathcal{M}_i \leftarrow \{m \in \mathcal{M} \mid s_i \in \textsc{Skills}(m) \wedge \textsc{Acc}(m) \geq \alpha\}$,  \textbf{for} $i = 1,\ldots,k$
  \State \textbf{if} $\exists i:\mathcal{M}_i=\varnothing$ \textbf{then continue} \Comment{no admissible model: skip workflow}
  {\setlength{\fboxsep}{1pt} \Statex \Comment{----------------------------------------\fbox{\textit{Stage 2: Pruning by operational cost}}---------------------------------------}}
  \State $\beta \leftarrow \sum_{i=1}^{k} \min_{m \in \mathcal{M}_i}\textsc{Size}(m)$; \quad $\textit{candidates} \leftarrow [\,]$ \Comment{parameter budget}
  \For{each $\mathbf{m}=(m_1,\ldots,m_k) \in \mathcal{M}_1 \times \cdots \times \mathcal{M}_k$}
    \State $active_s \leftarrow \textsc{ComputeActiveSize}(\mathbf{m})$; \textbf{if} $active_s \leq \beta$ \textbf{then} $\textit{candidates}.\textsc{append}((\mathbf{m},active_s))$
  \EndFor
  {\setlength{\fboxsep}{1pt} \Statex \Comment{---------------------------------------\fbox{\textit{Stage 3: Pruning by system constraints}}--------------------------------------}}
  \For{each $(\mathbf{m},active_s)$ in $\textit{candidates}$ sorted by $active_s$ ascending}
    \State $\mathcal{C}_{m_i} \leftarrow \{c \in \mathcal{C} \mid c \text{ is valid for } m_i\}, \; \forall m_i \in \mathbf{m}$; \quad $\mathcal{WV}_{\mathbf{m}} \leftarrow \varnothing$
    \For{each $(c_1,\ldots,c_k) \in \mathcal{C}_{m_1} \times \cdots \times \mathcal{C}_{m_k}$}
      \State $\mathbf{n} \leftarrow [\textsc{Node}(m_i,c_i)]_{i=1}^{k}$
      \State \textbf{if} $\sum_i \textsc{Lat}(n_i) > \lambda$ \textbf{then continue} \Comment{latency SLO}
      \State \textbf{if} $\exists i: \textsc{Cap}(n_i) < \textsc{Load}(n_i)+\sigma_i\rho$ \textbf{then continue} \Comment{capacity}
      \State $\textit{ok},\mathbf{gpu} \leftarrow \textsc{LeastLoaded}(\mathbf{n})$; \quad \textbf{if} $\neg\textit{ok}$ \textbf{then continue} \Comment{placement}
      \State $\mathcal{WV} \leftarrow (\mathbf{n},\mathbf{gpu})$
      \State \textbf{if} $\textsc{Cost}(\mathcal{WV}) < \textsc{Cost}(\mathcal{WV}_{\mathbf{m}})$ \textbf{then} $\mathcal{WV}_{\mathbf{m}}\leftarrow\mathcal{WV}$
    \EndFor
     \State \textbf{if} $\mathcal{WV}_{\mathbf{m}} \neq \varnothing$ \textbf{then} $\mathcal{WV}^{*} \leftarrow \arg\min_{\textsc{Cost}}(\mathcal{WV}^{*}, \mathcal{WV}_{\mathbf{m}})$; \textbf{break and go to line 2} \Comment{first feasible}
  \EndFor
\EndFor
\State \Return $\mathcal{WV}^{*}$
\end{algorithmic}
\end{algorithm}

\vspace{-1mm}
\noindent\textbf{Stage 1: Pruning by accuracy.}
For each workflow stage, {\system} retains only models that support the required skill and satisfy the accuracy SLO. Under our minimum-stage accuracy estimator, any assignment containing a model below $\alpha$ is infeasible. If no model can serve a stage, {\system} skips the workflow. Because this check is independent of deployment configuration, it eliminates model--configuration combinations before constructing them.

\vspace{-1mm}
\noindent\textbf{Stage 2: Pruning by operational cost.}
After accuracy pruning, {\system} constructs candidate model assignments. Evaluating every deployment configuration at this point would remain expensive, so it uses model size as a proxy for deployment cost: model memory and latency generally increase with parameter count~\citep{MaverIQ}. The parameter budget $\beta$ is the aggregate size of the smallest accuracy-feasible model for each stage, assuming no reuse. For an assignment $\mathbf{m}$, \textsc{ComputeActiveSize} counts the parameters of the distinct models that must be newly deployed after accounting for existing deployments and reuse within the workflow. {\system} retains assignments whose active size does not exceed $\beta$ and considers them in ascending active-size order. This heuristic favors assignments that exploit reuse or require a small incremental deployment footprint, but does not guarantee that every potentially lower-cost assignment is retained.

\vspace{-1mm}
\noindent\textbf{Stage 3: Pruning by system constraints.}
For each retained model assignment, in ascending active-size order, {\system} enumerates its valid tensor- and pipeline-parallelism configurations. It discards configurations that violate the end-to-end latency SLO or would cause a model instance to exceed its serving capacity. It then invokes \textsc{LeastLoaded}, which attempts to map the required model shards to GPUs with sufficient free memory while preferring less-loaded GPUs.

{\system} keeps each assignment's lowest-cost feasible configuration. It compares the first feasible assignment with the current best selection and proceeds to the next workflow. Thus, it selects the smallest-footprint feasible assignment per workflow and returns the lowest-cost selection across workflows.

\vspace{-1mm}
\noindent\textbf{Search cost.}
The worst-case search space remains exponential in the number of workflow stages. For $|\mathcal{W}|$ workflows with $|\mathcal{S}|$ stages, exhaustive selection considers
$O\left(
|\mathcal{W}| \cdot
(|\mathcal{M}| \cdot |\mathcal{C}| \cdot |\mathcal{D}|)^{|\mathcal{S}|}
\right)$
model--configuration assignments, in addition to placement checks (see Appendix~\ref{sec:appendix_algo} for details).

\vspace{-4mm}
\section{Implementation and Experimental Setup}
\label{sec:impl_eval}
\vspace{-4mm}

\noindent\textbf{Prototype and testbed.}
We implement {\system} in $\sim$3K lines of Python atop TensorRT-LLM~\citep{TensorRT-LLM}, with a centralized CPU controller and per-worker daemons reporting model load and free GPU memory; the Profiler uses MaverIQ~\citep{MaverIQ}. Experiments run on $8\times$NVIDIA RTX A6000 GPUs (48GB each) with a 64-core AMD EPYC 7763 CPU, and all systems execute inference through the same backend (see Appendix~\ref{app:implementation_details} for details).

\vspace{-1mm}
\noindent\textbf{Workloads and models.}
We evaluate seven generative-AI tasks~\citep{genAI_app_1, genAI_app_2, genAI_app_3}, each with three candidate workflows of 1--3 stages, using inputs from six public datasets~\citep{cnn_daily_mail_1, cnn_daily_mail_2, humaneval, agnews_yelp, dialogsum, writingprompts} and a catalog of 11 open-weight LLMs (1B--72B) from the Llama, Qwen, and Mistral families~\citep{llama-3, qwen-2.5, mistral}. Following prior LLM-serving evaluations~\citep{MaverIQ, DistServe, AlpaServe}, we replay a bursty code and a sustained conversation Azure trace~\citep{Splitwise} at four offered-load levels (25--100\%), preserving each trace's arrival structure and yielding 212--484 requests per run.
Appendix~\ref{app:experimental_details} provides more information regarding the experimental details.

\vspace{-1mm}
\noindent\textbf{Comparison methods.}
The baselines differ in which decisions they fix. \emph{LangChain}~\citep{LangChain} fixes all three, receiving a default workflow, the smallest proxy-accuracy-feasible models, and TP${}=1$, PP${}=1$. \emph{RouteLLM}~\citep{RouteLLM} frees model choice within a fixed workflow, routing each request between Qwen-2.5-14B and Qwen-2.5-7B at thresholds favoring maximum quality, minimum cost, or balance the two; we provision enough replicas to avoid queuing, intentionally favoring its latency, and count all replicas in its cost. \emph{Murakkab}~\citep{Murakkab} also frees deployment configurations but fixes the workflow and decides reuse offline; as its implementation is unavailable, we reconstruct its optimizer and run its selections on our backend, so our comparison concerns this configuration rather than its broader feature set. We also evaluate \emph{{\FMNoReuse}}, which applies {\system}'s selector without reuse, and compare the selector against brute force and Aquatope-style Bayesian optimization~\citep{AQUATOPE} (see Appendix~\ref{app:experimental_details} for more details).

\vspace{-1mm}
\noindent\textbf{SLOs and metrics.}
To set latency SLOs, for each task we measure its latency in isolation and set \textsc{Strict}--\textsc{Relaxed} latency SLOs as $2$--$5\times$ this value, following prior work~\citep{DistServe, Proteus, MaverIQ, AlpaServe}. 
To set accuracy SLOs, we define a model's proxy accuracy as its average score across six Open LLM Leaderboard benchmarks~\citep{OpenLLM_leaderboard}, and a variant's score as the minimum across its models. 
Following Murakkab, \textsc{Best}, \textsc{Good}, and \textsc{Basic} accuracy SLOs correspond to the p95, p80, and p50 achievable variant scores. 
Because we do not evaluate generated outputs, all accuracy compliance refers to this proxy. 
We report task latency, serving cost $\sum_j M_jT_jr$ (allocated GPU memory $\times$ residency time $\times$ GB-second rental rate), and cluster goodput: the rate of requests meeting their latency SLO under a variant that satisfies their proxy-accuracy SLO. Additional details appear in Appendix~\ref{app:experimental_details}.

\vspace{-4mm}
\section{Evaluation}
\label{sec:evaluation}
\vspace{-4mm}

We evaluate {\system}'s (1) end-to-end serving performance (\S~\ref{sec:eval_e2e}), (2) gains from joint workflow-variant selection and on-the-fly reuse (\S~\ref{sec:eval_js-otfr}), and (3) selection quality, efficiency, and scalability relative to brute-force search and Aquatope that leverages Bayesian optimization (\S~\ref{sec:eval_algo-efficiency}). Appendix~\ref{app:extra_eval} reports fixed-capacity scaling, SLO sensitivity, and system overheads.

\vspace{-4mm}
\subsection{End-to-End Serving Performance}
\label{sec:eval_e2e}
\vspace{-4mm}

\begin{figure*}[t]
\centering
\includegraphics[width=\linewidth]{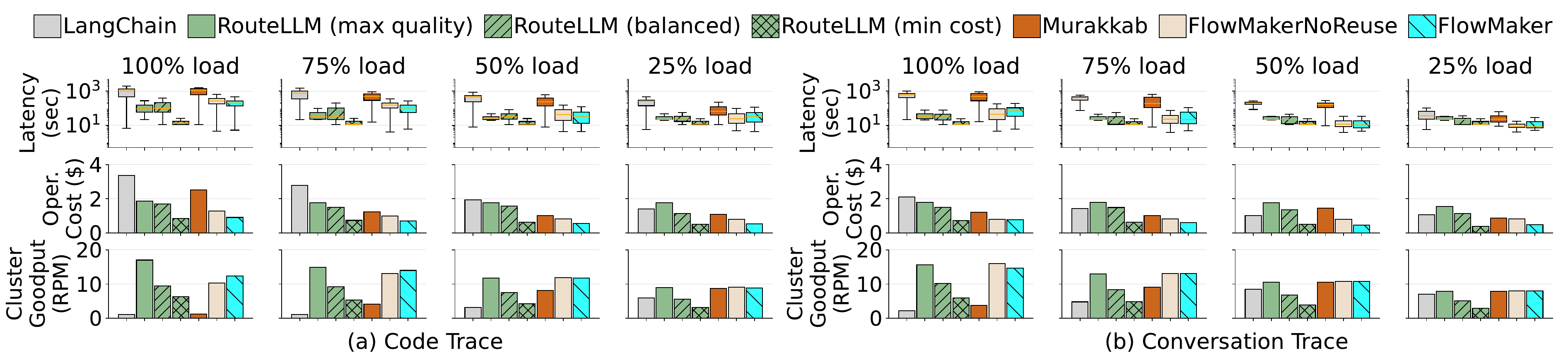}
\vspace{-8mm}
\caption{
End-to-end request latency (top), serving cost (middle), and cluster goodput (bottom) for 7 tasks under the Azure (a) code and (b) conversation traces. See \S~\ref{sec:eval_e2e} and \S~\ref{sec:eval_js-otfr}.
}
\label{fig:eval_e2e}
\vspace{-6mm}
\end{figure*}

Figure~\ref{fig:eval_e2e} compares the systems on the same seven-task workload and task-specific SLOs, under bursty code and sustained conversation arrivals across four load levels (see Appendix~\ref{app:e2e_detailed} for additional details). 
The systems differ in which of three decisions they can adapt: the workflow, the models used, and their deployment configurations. Fixing any of these decisions limits adaptation to workload and cluster conditions, reducing efficiency or SLO compliance.

LangChain fixes all three decisions and never reuses models. It deploys 15 instances (290GB) for 6 tasks but cannot place the 7\textsuperscript{th} because memory fragmentation leaves no GPU with enough free capacity. 
At 50\% code-trace load, it colocates 4 instances on one GPU, increasing average latency to 426s versus 38s for {\system}, which serves all 7 tasks with only 3--5 instances. 
Overall, {\system} achieves $1.13$--$12.84\times$ higher goodput ($5.2\times$ avg) at $2.2$--$4\times$ lower cost ($2.93\times$ avg).

RouteLLM can choose between models, but fixes the workflow and restricts deployment to a pre-deployed strong/weak pair. 
Its routing threshold therefore trades quality for cost only by shifting requests between these two models. 
At maximum quality, it meets all proxy-accuracy SLOs and achieves on average 5.8\% higher goodput than {\system}, but reserves 206--235GB and costs $2.06$--$3.84\times$ ($2.95\times$ avg) more. Its minimum-cost and balanced settings meet only 34.9--36.8\% and 62.6--65.6\% of proxy-accuracy SLOs, versus 100\% for {\system}, yielding $2.62\times$ and $1.52\times$ lower average goodput. 
RouteLLM's lower latency comes from provisioning enough replicas to avoid queuing, which also drives up cost. More details are in Appendix~\ref{app:e2e_detailed}.

Murakkab additionally adapts deployment configurations, but still fixes the workflow and chooses reuse offline. 
At low load, goodput is similar (up to $1.46\times$, $1.21\times$ avg), while {\system} is $1.84$--$3.15\times$ ($2.23\times$ avg) cheaper by selecting shorter workflows when beneficial. 
At high load, Murakkab's fixed workflow and reuse choices increase latency and queuing; {\system} achieves $1.45$--$10.14\times$ ($4.73\times$ avg) higher goodput at $1.56$--$2.77\times$ ($1.95\times$ avg) lower cost.

\vspace{-2mm}
\noindent
\begingroup
\setlength{\fboxsep}{4pt}
\setlength{\fboxrule}{0.05mm}
\hspace{-2.5mm}
\makebox[\linewidth][l]{%
\fcolorbox{black}{gray!10}{%
\parbox{\dimexpr\linewidth-2\fboxsep-2\fboxrule\relax}{%
\textbf{End-to-end takeaway.}
At high load, {\system} achieves $1.45$--$10.14\times$ Murakkab's goodput ($4.73\times$ on average) while reducing serving cost by $1.56$--$2.77\times$ ($1.95\times$ on average). RouteLLM lowers latency through replication, but its maximum-quality setting costs $2.06$--$3.84\times$ more, while only 34.9--36.8\% of requests satisfy their proxy-accuracy SLOs in its minimum-cost setting.
}%
}%
}%
\endgroup

\vspace{-3mm}
\subsection{Impact of Joint Selection and On-the-Fly Reuse}
\label{sec:eval_js-otfr}
\vspace{-3mm}

\noindent\textbf{Joint selection across workflows, models, and deployment configurations.}
We compare {\FMNoReuse}, which jointly selects across all three dimensions but deploys separate instances per task, with Murakkab, which fixes each task's workflow but retains offline reuse (Figure~\ref{fig:eval_e2e}). Because the systems also differ in reuse policy, this is conservative evidence for joint selection rather than a controlled isolation: {\FMNoReuse} operates without reuse, while Murakkab benefits from it. 
At lower loads, {\FMNoReuse} achieves up to $1.47\times$ ($1.13\times$ on average) higher goodput and reduces cost by up to $1.83\times$ ($1.38\times$ on average). Murakkab consumes less memory through reuse---117.11GB rather than 186.46GB in one conversation-trace setting---but its fixed workflows contain at least two stages, whereas {\FMNoReuse} can select one- or two-stage alternatives. Across tasks and loads, these choices increase Murakkab's latency by $2.41$--$10.73\times$, offsetting its memory savings.
At higher loads, the additional latency causes queues to grow. Across both traces, {\FMNoReuse} achieves $1.45$--$8.46\times$ higher goodput ($4.34\times$ on average) and reduces cost by $1.24$--$1.95\times$ ($1.48\times$ on average).

\noindent\textbf{On-the-fly reuse.}
We isolate reuse by comparing {\system} with {\FMNoReuse}; both apply the same joint-selection heuristic, but only {\system} can map stages from different tasks to shared instances. Their goodput remains similar across loads. At lower loads, {\system} reuses 3--4 instances, whereas {\FMNoReuse} deploys 8, reducing cost by $1.47$--$1.73\times$ ($1.61\times$ avg). At higher loads, {\system} shares less aggressively and deploys five instances to preserve capacity, but still reduces cost by $1.04$--$1.42\times$ ($1.31\times$ avg).
For example, under the code trace at 100\% load, {\system} selects a three-stage variant that reuses an existing Qwen-2.5-3B instance, whereas {\FMNoReuse} selects a two-stage variant requiring new deployments. Thus, the lower-cost choice depends on both the workflow structure and the current deployment state.

\vspace{-1mm}
\noindent
\begingroup
\setlength{\fboxsep}{4pt}
\setlength{\fboxrule}{0.05mm}
\hspace{-2.5mm}
\makebox[\linewidth][l]{%
\fcolorbox{black}{gray!10}{%
\parbox{\dimexpr\linewidth-2\fboxsep-2\fboxrule\relax}{%
\textbf{Ablation takeaway.}
Despite disabling reuse, {\FMNoReuse} achieves $1.45$--$8.46\times$ ($4.34\times$ on average) higher goodput and $1.24$--$1.95\times$ ($1.48\times$ on average) lower cost than Murakkab at high load. Adding load-aware reuse further reduces {\system}'s cost by $1.04$--$1.42\times$ ($1.31\times$ on average) without materially changing goodput.
}%
}%
}%
\endgroup

\vspace{-3mm}
\subsection{Selection Quality, Efficiency, and Scalability}
\label{sec:eval_algo-efficiency}
\vspace{-3mm}

\begin{wrapfigure}{r}{0.4\linewidth}
\centering
\vspace{-6mm}
\includegraphics[width=\linewidth]{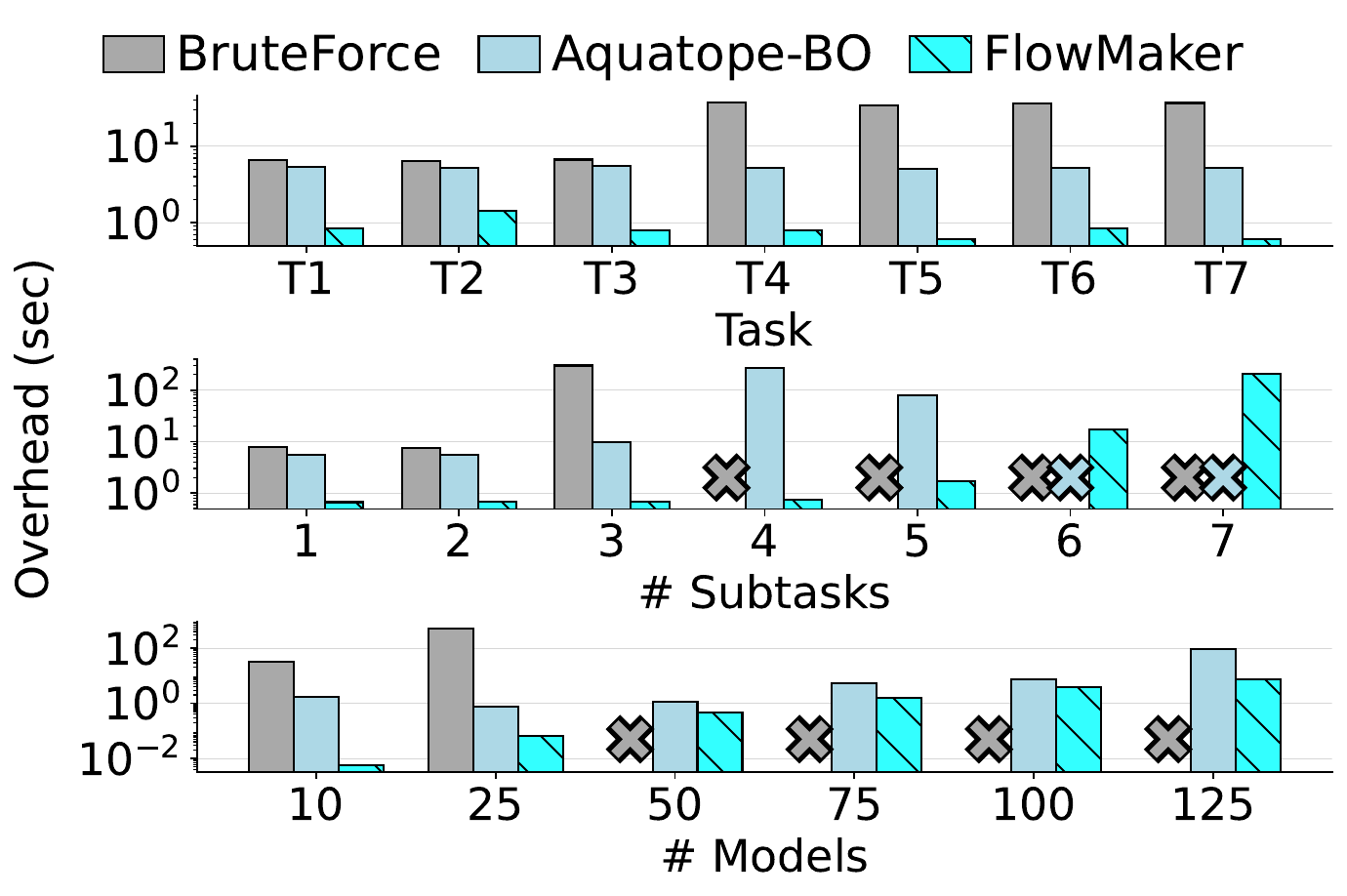}
\vspace{-8mm}
\caption{\small{
Workflow-variant selection time across tasks (top), workflow depth (middle), and model-catalog size (bottom).
}}
\label{fig:eval_algoeff}
\vspace{-6mm}
\end{wrapfigure}
We compare {\system} with brute-force (BF) and Aquatope-style Bayesian optimization (BO)~\citep{BO_org, BO_textbook, AQUATOPE, Composer}. BF enumerates the candidate space and returns the minimum-cost feasible variant under Equation~\ref{eq:cost}. BO uses Optuna's TPE sampler~\citep{optuna_tpe_sampler}, performs 200 sequential trials, and penalizes SLO violations.

\vspace{-1mm}
\noindent\textbf{Selection quality.}
Across the seven tasks and every comparable point in the model-catalog and workflow-depth scaling experiments, {\system}, BF, and BO select the same workflow variant. Thus, {\system} has a 100\% observed match rate, a 0\% observed objective gap relative to BF, and the same feasibility and SLO-satisfaction outcome as both reference methods. 
This provides empirical evidence that {\system}'s algorithm matches the selections of BF without incurring its substantial overheads.

\vspace{-1mm}
\noindent\textbf{Time efficiency.}
Figure~\ref{fig:eval_algoeff} (top) shows that {\system} is $3.7$--$8.5\times$ faster than BO and $4.5$--$61\times$ faster than BF across the seven tasks. BF takes approximately 6.5s for the simpler tasks (T1--T3) and approximately 40s for the larger candidate spaces of T4--T7. BO requires approximately 5s per task for its 200 sequential trials. In contrast, {\system}'s staged heuristic filters candidates using proxy accuracy and active deployment size before evaluating deployment configurations, selecting a variant in approximately one second.
The approximately 40s reported here measures BF enumeration over already-computed model profiles. In contrast, the approximately 40 CPU-minutes reported in \S~\ref{sec:full-search-space} measure end-to-end exhaustive evaluation of the example task, including computing the latency and memory estimates required to evaluate its 24,310 variants.

\vspace{-1mm}
\noindent\textbf{Scalability.}
Workflow depth most strongly affects search time (Figure~\ref{fig:eval_algoeff}, middle). 
BF's and BO's search time exceeds 15 minutes at $\geq6$ stages (indicated by \texttt{x} in the Figure), while {\system} stays below two seconds for 3--5 stages, yielding $14.5$--$443\times$ speedups. 
At 7 stages, however, its search time rises to 210s, reflecting the underlying combinatorial growth. 
{\system} also scales better with model-catalog size (bottom): BF times out at 50 models, BO slows by $12.2\times$ at 125 models, while {\system}'s early filtering reduces the model--configuration combinations explored.

\vspace{-5mm}
\section{Conclusion}
\label{sec:conclusion}
\vspace{-4mm}

{\system} is a serving system that takes a task and its latency and accuracy SLOs and jointly selects a workflow variant and state-aware reuse plan across workflows, models, and deployment configurations. Its constraint-guided selector matches brute-force and Bayesian-optimization selections  while reducing selection time by $4.5$--$61\times$ and $3.7$--$8.5\times$, respectively. At high load, {\system} achieves $4.73\times$ Murakkab's goodput while reducing serving cost by $1.95\times$ on average, showing the benefits of joint, state-aware selection for cost-efficient generative AI workflow serving.

\section*{AI Use Statement}
In this work, we used generative AI tools to help implement and tune the baseline systems we compare against, to write and edit software code for our prototype and experiment scripts, and to improve the readability of parts of the manuscript.
We have not used generative AI tools to develop the conceptual framework, formulate claims, design the methodology or experiments, or interpret results, and the remaining categories in the ICLR policy are not applicable. We have reviewed all AI-assisted work: AI-assisted code was verified and tested by the authors, and baseline implementations were checked against their original papers. We take responsibility for the final content of this work, including text, claims, and artifacts produced with the aid of generative AI.


\bibliography{iclr2027_conference}

@inproceedings{AsyncHow,
author = {Lin, Fangru and Malfa, Emanuele La and Hofmann, Valentin and Yang, Elle Michelle and Cohn, Anthony G. and Pierrehumbert, Janet B.},
title = {{Graph-enhanced large language models in asynchronous plan reasoning}},
year = {2024},
publisher = {JMLR.org},
booktitle = {Proceedings of the 41st International Conference on Machine Learning},
articleno = {1213},
numpages = {27},
location = {Vienna, Austria},
series = {ICML'24},
url={https://dl.acm.org/doi/10.5555/3692070.3693283}
}

@inproceedings {Murakkab,
author = {Gohar Irfan Chaudhry and Esha Choukse and Haoran Qiu and Inigo Goiri and Rodrigo Fonseca and Adam Belay and Ricardo Bianchini},
title = {{Murakkab: {Resource-Efficient} Agentic Workflow Orchestration in Cloud Platforms}},
booktitle = {20th USENIX Symposium on Operating Systems Design and Implementation (OSDI 26)},
year = {2026},
isbn = {978-1-939133-55-7},
address = {Seattle, WA},
pages = {567--587},
url = {https://www.usenix.org/conference/osdi26/presentation/chaudhry},
publisher = {USENIX Association},
month = jul
}

@misc{LangChain,
  title        = {{LangChain}},
  author       = {Chase, Harrison},
  year         = {2022},
  howpublished = {\url{https://www.langchain.com/}},
}

@misc{LangGraph,
  author       = {{LangChain}},
  title        = {{LangGraph: Agent Orchestration Framework for Reliable AI Applications}},
  year         = {2025},
  howpublished = {\url{https://www.langchain.com/langgraph}},
}

@misc{AWS_Quick_Flows,
  author       = {{Amazon Web Services}},
  title        = {{Amazon Quick Flows}},
  year         = {2025},
  howpublished = {\url{https://aws.amazon.com/quick/flows/}},
}

@misc{Azure_AI_foundry,
  author       = {{Microsoft Azure}},
  title        = {{Azure AI Foundry: Build, Deploy, and Manage AI Applications and Agents}},
  year         = {2025},
  howpublished = {\url{https://azure.microsoft.com/en-us/products/ai-foundry/}},
}

@misc{CrewAI,
  author       = {{CrewAI}},
  title        = {{CrewAI: Multi-Agent Platform for Automating AI Workflows}},
  year         = {2024},
  url          = {https://crewai.com/},
}

@inproceedings{MaverIQ,
author = {Liakopoulos, Dimitrios and Sinha, Prasoon and Hu, Tianrui and Lee, Myungjin and Yadwadkar, Neeraja J.},
title = {{MaverIQ: Fingerprint-Guided Extrapolation and Fragmentation-Aware Layering for Intent-Based LLM Serving}},
year = {2025},
isbn = {9798400714665},
publisher = {Association for Computing Machinery},
address = {New York, NY, USA},
url = {https://doi.org/10.1145/3712285.3759867},
doi = {10.1145/3712285.3759867},
booktitle = {Proceedings of the International Conference for High Performance Computing, Networking, Storage and Analysis},
pages = {1676–1696},
numpages = {21},
location = {
},
series = {SC '25}
}

@inproceedings{Vidur,
 author = {Agrawal, Amey and Kedia, Nitin and Mohan, Jayashree and Panwar, Ashish and Kwatra, Nipun and Gulavani, Bhargav S. and Ramjee, Ramachandran and Tumanov, Alexey},
 booktitle = {Proceedings of Machine Learning and Systems},
 editor = {P. Gibbons and G. Pekhimenko and C. De Sa},
 pages = {351--366},
 title = {{VIDUR: A LARGE-SCALE SIMULATION FRAMEWORK FOR LLM INFERENCE}},
 url = {https://proceedings.mlsys.org/paper_files/paper/2024/file/b74a8de47d2b3c928360e0a011f48351-Paper-Conference.pdf},
 volume = {6},
 year = {2024}
}

@inproceedings{MAD_Max,
author = {Hsia, Samuel and Golden, Alicia and Acun, Bilge and Ardalani, Newsha and DeVito, Zachary and Wei, Gu-Yeon and Brooks, David and Wu, Carole-Jean},
title = {{MAD-Max Beyond Single-Node: Enabling Large Machine Learning Model Acceleration on Distributed Systems}},
year = {2025},
isbn = {9798350326581},
publisher = {IEEE Press},
url = {https://doi.org/10.1109/ISCA59077.2024.00064},
doi = {10.1109/ISCA59077.2024.00064},
booktitle = {Proceedings of the 51st Annual International Symposium on Computer Architecture},
pages = {818–833},
numpages = {16},
location = {Buenos Aires, Argentina},
series = {ISCA '24}
}

@inproceedings{NeuSight,
author = {Lee, Seonho and Phanishayee, Amar and Mahajan, Divya},
title = {{Forecasting GPU Performance for Deep Learning Training and Inference}},
year = {2025},
isbn = {9798400706981},
publisher = {Association for Computing Machinery},
address = {New York, NY, USA},
url = {https://doi.org/10.1145/3669940.3707265},
doi = {10.1145/3669940.3707265},
booktitle = {Proceedings of the 30th ACM International Conference on Architectural Support for Programming Languages and Operating Systems, Volume 1},
pages = {493–508},
numpages = {16},
location = {Rotterdam, Netherlands},
series = {ASPLOS '25}
}

@inproceedings{BurstGPT,
author = {Wang, Yuxin and Chen, Yuhan and Li, Zeyu and Kang, Xueze and Fang, Yuchu and Zhou, Yeju and Zheng, Yang and Tang, Zhenheng and He, Xin and Guo, Rui and Wang, Xin and Wang, Qiang and Zhou, Amelie Chi and Chu, Xiaowen},
title = {{BurstGPT: A Real-World Workload Dataset to Optimize LLM Serving Systems}},
year = {2025},
isbn = {9798400714542},
publisher = {Association for Computing Machinery},
address = {New York, NY, USA},
url = {https://doi.org/10.1145/3711896.3737413},
doi = {10.1145/3711896.3737413},
booktitle = {Proceedings of the 31st ACM SIGKDD Conference on Knowledge Discovery and Data Mining V.2},
pages = {5831–5841},
numpages = {11},
location = {Toronto ON, Canada},
series = {KDD '25}
}

@inproceedings{Splitwise,
author = {Patel, Pratyush and Choukse, Esha and Zhang, Chaojie and Shah, Aashaka and Goiri, Íñigo and Maleki, Saeed and Bianchini, Ricardo},
title = {{Splitwise: Efficient generative LLM inference using phase splitting}},
booktitle = {ISCA},
year = {2024},
month = {June},
url = {https://www.microsoft.com/en-us/research/publication/splitwise-efficient-generative-llm-inference-using-phase-splitting/},
}

@INPROCEEDINGS{DynamoLLM,
  author={Stojkovic, Jovan and Zhang, Chaojie and Goiri, Íñigo and Torrellas, Josep and Choukse, Esha},
  booktitle={2025 IEEE International Symposium on High Performance Computer Architecture (HPCA)}, 
  title={{DynamoLLM: Designing LLM Inference Clusters for Performance and Energy Efficiency}}, 
  year={2025},
  volume={},
  number={},
  pages={1348-1362},
  doi={10.1109/HPCA61900.2025.00102},
  url={https://doi.org/10.1109/HPCA61900.2025.00102}}

@inproceedings{ModServe,
author = {Qiu, Haoran and Biswas, Anish and Zhao, Zihan and Mohan, Jayashree and Khare, Alind and Choukse, Esha and Goiri, \'{I}\~{n}igo and Zhang, Zeyu and Shen, Haiying and Bansal, Chetan and Ramjee, Ram and Fonseca, Rodrigo},
title = {{ModServe: Modality- and Stage-Aware Resource Disaggregation for Scalable Multimodal Model Serving}},
year = {2026},
isbn = {9798400722769},
publisher = {Association for Computing Machinery},
address = {New York, NY, USA},
url = {https://doi.org/10.1145/3772052.3772254},
doi = {10.1145/3772052.3772254},
booktitle = {Proceedings of the 2025 ACM Symposium on Cloud Computing},
pages = {817–830},
numpages = {14},
location = {
},
series = {SoCC '25}
}

@misc{OpenLLM_leaderboard,
  author       = {Clémentine Fourrier and Nathan Habib and Alina Lozovskaya and Konrad Szafer and Thomas Wolf},
  title        = {{Open LLM Leaderboard}},
  year         = {2024},
  publisher    = {Hugging Face},
  howpublished = {\url{https://huggingface.co/open-llm-leaderboard}},
  note         = {Accessed: 2026-03-18}
}

@inproceedings{AFlow,
title={{{AF}low: Automating Agentic Workflow Generation}},
author={Jiayi Zhang and Jinyu Xiang and Zhaoyang Yu and Fengwei Teng and Xiong-Hui Chen and Jiaqi Chen and Mingchen Zhuge and Xin Cheng and Sirui Hong and Jinlin Wang and Bingnan Zheng and Bang Liu and Yuyu Luo and Chenglin Wu},
booktitle={The Thirteenth International Conference on Learning Representations},
year={2025},
url={https://openreview.net/forum?id=z5uVAKwmjf}
}

@inproceedings{ADAS,
title={{Automated Design of Agentic Systems}},
author={Shengran Hu and Cong Lu and Jeff Clune},
booktitle={The Thirteenth International Conference on Learning Representations},
year={2025},
url={https://openreview.net/forum?id=t9U3LW7JVX}
}

@misc{BAIR_CompoundAI,
  title={{The Shift from Models to Compound AI Systems}},
  author={Matei Zaharia and Omar Khattab and Lingjiao Chen and Jared Quincy Davis
          and Heather Miller and Chris Potts and James Zou and Michael Carbin
          and Jonathan Frankle and Naveen Rao and Ali Ghodsi},
  howpublished={\url{https://bair.berkeley.edu/blog/2024/02/18/compound-ai-systems/}},
  year={2024}
}

@misc{AutoFlow,
      title={{AutoFlow: Automated Workflow Generation for Large Language Model Agents}}, 
      author={Zelong Li and Shuyuan Xu and Kai Mei and Wenyue Hua and Balaji Rama and Om Raheja and Hao Wang and He Zhu and Yongfeng Zhang},
      year={2024},
      eprint={2407.12821},
      archivePrefix={arXiv},
      primaryClass={cs.CL},
      url={https://arxiv.org/abs/2407.12821}, 
}

@inproceedings{AutoGen,
title={{AutoGen: Enabling Next-Gen {LLM} Applications via Multi-Agent Conversations}},
author={Qingyun Wu and Gagan Bansal and Jieyu Zhang and Yiran Wu and Beibin Li and Erkang Zhu and Li Jiang and Xiaoyun Zhang and Shaokun Zhang and Jiale Liu and Ahmed Hassan Awadallah and Ryen W White and Doug Burger and Chi Wang},
booktitle={First Conference on Language Modeling},
year={2024},
url={https://openreview.net/forum?id=BAakY1hNKS}
}

@misc{Nalar,
      title={Nalar: An agent serving framework}, 
      author={Marco Laju and Donghyun Son and Saurabh Agarwal and Nitin Kedia and Myungjin Lee and Jayanth Srinivasa and Aditya Akella},
      year={2026},
      eprint={2601.05109},
      archivePrefix={arXiv},
      primaryClass={cs.DC},
      url={https://arxiv.org/abs/2601.05109}, 
}

@misc{Helium,
      title={{Efficient LLM Serving for Agentic Workflows: A Data Systems Perspective}}, 
      author={Noppanat Wadlom and Junyi Shen and Yao Lu},
      year={2026},
      eprint={2603.16104},
      archivePrefix={arXiv},
      primaryClass={cs.MA},
      url={https://arxiv.org/abs/2603.16104}, 
}

@inproceedings {Parrot,
author = {Chaofan Lin and Zhenhua Han and Chengruidong Zhang and Yuqing Yang and Fan Yang and Chen Chen and Lili Qiu},
title = {{Parrot: Efficient Serving of {LLM-based} Applications with Semantic Variable}},
booktitle = {18th USENIX Symposium on Operating Systems Design and Implementation (OSDI 24)},
year = {2024},
isbn = {978-1-939133-40-3},
address = {Santa Clara, CA},
pages = {929--945},
url = {https://www.usenix.org/conference/osdi24/presentation/lin-chaofan},
publisher = {USENIX Association},
month = jul
}

@inproceedings{MetaGPT,
title={{Meta{GPT}: Meta Programming for A Multi-Agent Collaborative Framework}},
author={Sirui Hong and Mingchen Zhuge and Jonathan Chen and Xiawu Zheng and Yuheng Cheng and Jinlin Wang and Ceyao Zhang and Zili Wang and Steven Ka Shing Yau and Zijuan Lin and Liyang Zhou and Chenyu Ran and Lingfeng Xiao and Chenglin Wu and J{\"u}rgen Schmidhuber},
booktitle={The Twelfth International Conference on Learning Representations},
year={2024},
url={https://openreview.net/forum?id=VtmBAGCN7o}
}

@inproceedings{KVFlow,
title={{KVF}low: Efficient Prefix Caching for Accelerating {LLM}-Based Multi-Agent Workflows},
author={Zaifeng Pan and AJJKUMAR PATEL and Yipeng Shen and Zhengding Hu and Yue Guan and Wan-Lu Li and Lianhui Qin and Yida Wang and Yufei Ding},
booktitle={The Thirty-ninth Annual Conference on Neural Information Processing Systems},
year={2025},
url={https://openreview.net/forum?id=5Iw1nDtYmT}
}

@inproceedings{AIOS,
title={{AIOS}: {LLM} Agent Operating System},
author={Kai Mei and Xi Zhu and Wujiang Xu and Mingyu Jin and Wenyue Hua and Zelong Li and Shuyuan Xu and Ruosong Ye and Yingqiang Ge and Yongfeng Zhang},
booktitle={Second Conference on Language Modeling},
year={2025},
url={https://openreview.net/forum?id=L4HHkCDz2x}
}

@misc{Aragog,
      title={Aragog: Just-in-Time Model Routing for Scalable Serving of Agentic Workflows}, 
      author={Yinwei Dai and Zhuofu Chen and Anand Iyer and Ravi Netravali},
      year={2025},
      eprint={2511.20975},
      archivePrefix={arXiv},
      primaryClass={cs.DC},
      url={https://arxiv.org/abs/2511.20975}, 
}

@misc{DAAO,
      title={{Difficulty-Aware Agentic Orchestration for Query-Specific Multi-Agent Workflows}}, 
      author={Jinwei Su and Qizhen Lan and Yinghui Xia and Lifan Sun and Weiyou Tian and Tianyu Shi and Xinyuan Song and Lewei He and Yang Jingsong},
      year={2026},
      eprint={2509.11079},
      archivePrefix={arXiv},
      primaryClass={cs.AI},
      url={https://arxiv.org/abs/2509.11079}, 
}

@inproceedings{Cognify,
author = {He, Zijian and Abhyankar, Reyna and Srivatsa, Vikranth and Zhang, Yiying},
title = {{Cognify: Supercharging Gen-AI Workflows With Hierarchical Autotuning}},
year = {2025},
isbn = {9798400714542},
publisher = {Association for Computing Machinery},
address = {New York, NY, USA},
url = {https://doi.org/10.1145/3711896.3736884},
doi = {10.1145/3711896.3736884},
booktitle = {Proceedings of the 31st ACM SIGKDD Conference on Knowledge Discovery and Data Mining V.2},
pages = {932–943},
numpages = {12},
location = {Toronto ON, Canada},
series = {KDD '25}
}

@misc{DSPy,
      title={{DSPy: Compiling Declarative Language Model Calls into Self-Improving Pipelines}}, 
      author={Omar Khattab and Arnav Singhvi and Paridhi Maheshwari and Zhiyuan Zhang and Keshav Santhanam and Sri Vardhamanan and Saiful Haq and Ashutosh Sharma and Thomas T. Joshi and Hanna Moazam and Heather Miller and Matei Zaharia and Christopher Potts},
      year={2023},
      eprint={2310.03714},
      archivePrefix={arXiv},
      primaryClass={cs.CL},
      url={https://arxiv.org/abs/2310.03714}, 
}

@inproceedings{LLMSelector,
title={{LLMSELECTOR}: Learning to Select Models in Compound {AI} Systems},
author={Lingjiao Chen and Jared Quincy Davis and Boris Hanin and Peter Bailis and James Zou and Matei Zaharia and Ion Stoica},
booktitle={ICML 2025 Workshop on Collaborative and Federated Agentic Workflows},
year={2025},
url={https://openreview.net/forum?id=NphowWHYJj}
}

@inproceedings{RouteLLM,
title={{Route{LLM}: Learning to Route {LLM}s from Preference Data}},
author={Isaac Ong and Amjad Almahairi and Vincent Wu and Wei-Lin Chiang and Tianhao Wu and Joseph E. Gonzalez and M Waleed Kadous and Ion Stoica},
booktitle={The Thirteenth International Conference on Learning Representations},
year={2025},
url={https://openreview.net/forum?id=8sSqNntaMr}
}

@inproceedings{MasRouter,
    title = "{M}as{R}outer: Learning to Route {LLM}s for Multi-Agent Systems",
    author = "Yue, Yanwei  and
      Zhang, Guibin  and
      Liu, Boyang  and
      Wan, Guancheng  and
      Wang, Kun  and
      Cheng, Dawei  and
      Qi, Yiyan",
    editor = "Che, Wanxiang  and
      Nabende, Joyce  and
      Shutova, Ekaterina  and
      Pilehvar, Mohammad Taher",
    booktitle = "Proceedings of the 63rd Annual Meeting of the Association for Computational Linguistics (Volume 1: Long Papers)",
    month = jul,
    year = "2025",
    address = "Vienna, Austria",
    publisher = "Association for Computational Linguistics",
    url = "https://aclanthology.org/2025.acl-long.757/",
    doi = "10.18653/v1/2025.acl-long.757",
    pages = "15549--15572",
    ISBN = "979-8-89176-251-0"
}

@inproceedings{vLLM,
author = {Kwon, Woosuk and Li, Zhuohan and Zhuang, Siyuan and Sheng, Ying and Zheng, Lianmin and Yu, Cody Hao and Gonzalez, Joseph and Zhang, Hao and Stoica, Ion},
title = {{Efficient Memory Management for Large Language Model Serving with PagedAttention}},
year = {2023},
isbn = {9798400702297},
publisher = {Association for Computing Machinery},
address = {New York, NY, USA},
url = {https://doi.org/10.1145/3600006.3613165},
doi = {10.1145/3600006.3613165},
booktitle = {Proceedings of the 29th Symposium on Operating Systems Principles},
pages = {611–626},
numpages = {16},
location = {Koblenz, Germany},
series = {SOSP '23}
}

@misc{TensorRT-LLM,
  author       = {{NVIDIA Corporation}},
  title        = {{TensorRT-LLM: Optimized Inference for Large Language Models}},
  year         = {2024},
  howpublished = {\url{https://github.com/NVIDIA/TensorRT-LLM}},
}

@inproceedings{DeepSpeed,
author = {Aminabadi, Reza Yazdani and Rajbhandari, Samyam and Awan, Ammar Ahmad and Li, Cheng and Li, Du and Zheng, Elton and Ruwase, Olatunji and Smith, Shaden and Zhang, Minjia and Rasley, Jeff and He, Yuxiong},
title = {{DeepSpeed-inference: enabling efficient inference of transformer models at unprecedented scale}},
year = {2022},
isbn = {9784665454445},
publisher = {IEEE Press},
booktitle = {Proceedings of the International Conference on High Performance Computing, Networking, Storage and Analysis},
articleno = {46},
numpages = {15},
location = {Dallas, Texas},
series = {SC '22},
url={https://dl.acm.org/doi/abs/10.5555/3571885.3571946}
}

@inproceedings {AlpaServe,
author = {Zhuohan Li and Lianmin Zheng and Yinmin Zhong and Vincent Liu and Ying Sheng and Xin Jin and Yanping Huang and Zhifeng Chen and Hao Zhang and Joseph E. Gonzalez and Ion Stoica},
title = {{AlpaServe}: Statistical Multiplexing with Model Parallelism for Deep Learning Serving},
booktitle = {17th USENIX Symposium on Operating Systems Design and Implementation (OSDI 23)},
year = {2023},
isbn = {978-1-939133-34-2},
address = {Boston, MA},
pages = {663--679},
url = {https://www.usenix.org/conference/osdi23/presentation/li-zhouhan},
publisher = {USENIX Association},
month = jul
}

@misc{ScaffoldedLLMs,
      title={{Scaffolded Language Models with Language Supervision for Mixed-Autonomy: A Survey}}, 
      author={Matthieu Lin and Jenny Sheng and Andrew Zhao and Shenzhi Wang and Yang Yue and Victor Shea Jay Huang and Huan Liu and Jun Liu and Gao Huang and Yong-Jin Liu},
      year={2025},
      eprint={2410.16392},
      archivePrefix={arXiv},
      primaryClass={cs.CL},
      url={https://arxiv.org/abs/2410.16392}, 
}

@misc{Google_Vertex_Agent_Engine,
  title        = {Vertex AI Agent Engine Overview},
  author       = {{Google Cloud}},
  year         = {2025},
  howpublished = {\url{https://docs.cloud.google.com/agent-builder/agent-engine/overview}},
}

@misc{openai_gpt_oss_120b_api,
  title        = {{gpt-oss-120b Model}},
  author       = {{OpenAI}},
  year         = {2025},
  howpublished = {\url{https://developers.openai.com/api/docs/models/gpt-oss-120b}},
  note         = {OpenAI API Documentation}
}

@inproceedings{CoT,
author = {Wei, Jason and Wang, Xuezhi and Schuurmans, Dale and Bosma, Maarten and Ichter, Brian and Xia, Fei and Chi, Ed H. and Le, Quoc V. and Zhou, Denny},
title = {{Chain-of-thought prompting elicits reasoning in large language models}},
year = {2022},
isbn = {9781713871088},
publisher = {Curran Associates Inc.},
address = {Red Hook, NY, USA},
booktitle = {Proceedings of the 36th International Conference on Neural Information Processing Systems},
articleno = {1800},
numpages = {14},
location = {New Orleans, LA, USA},
series = {NIPS '22},
url={https://dl.acm.org/doi/10.5555/3600270.3602070}
}

@misc{NVML,
  title        = {{NVIDIA Management Library (NVML)}},
  author       = {{NVIDIA Corporation}},
  year         = {2025},
  howpublished = {\url{https://developer.nvidia.com/management-library-nvml}},
  note         = {Accessed: 2026-03-31}
}

@inproceedings{Composer,
title={Composer: A Search Framework for Hybrid Neural Architecture Design},
author={Bilge Acun and Prasoon Sinha and Newsha Ardalani and Sangmin Bae and Alicia Golden and Meghana Madhyastha and Chien-Yu Lin and Fei Sun and Neeraja J. Yadwadkar and Carole-Jean Wu},
booktitle={The Fourteenth International Conference on Learning Representations},
year={2026},
url={https://openreview.net/forum?id=m00gjQfpCc}
}

@inproceedings{SWE-Agent,
 author = {Yang, John and Jimenez, Carlos E. and Wettig, Alexander and Lieret, Kilian and Yao, Shunyu and Narasimhan, Karthik and Press, Ofir},
 booktitle = {Advances in Neural Information Processing Systems},
 doi = {10.52202/079017-1601},
 editor = {A. Globerson and L. Mackey and D. Belgrave and A. Fan and U. Paquet and J. Tomczak and C. Zhang},
 pages = {50528--50652},
 publisher = {Curran Associates, Inc.},
 title = {{SWE-agent: Agent-Computer Interfaces Enable Automated Software Engineering}},
 url = {https://proceedings.neurips.cc/paper_files/paper/2024/file/5a7c947568c1b1328ccc5230172e1e7c-Paper-Conference.pdf},
 volume = {37},
 year = {2024}
}

@inproceedings{Proteus,
author = {Ahmad, Sohaib and Guan, Hui and Friedman, Brian D. and Williams, Thomas and Sitaraman, Ramesh K. and Woo, Thomas},
title = {{Proteus: A High-Throughput Inference-Serving System with Accuracy Scaling}},
year = {2024},
isbn = {9798400703720},
publisher = {Association for Computing Machinery},
address = {New York, NY, USA},
url = {https://doi.org/10.1145/3617232.3624849},
doi = {10.1145/3617232.3624849},
booktitle = {Proceedings of the 29th ACM International Conference on Architectural Support for Programming Languages and Operating Systems, Volume 1},
pages = {318–334},
numpages = {17},
location = {La Jolla, CA, USA},
series = {ASPLOS '24}
}

@misc{aws_lambda_pricing,
  title        = {{AWS Lambda Pricing}},
  author       = {{Amazon Web Services}},
  year         = {2026},
  howpublished = {\url{https://aws.amazon.com/lambda/pricing/}},
}

@inproceedings{Serveless_Cost,
author = {Eismann, Simon and Grohmann, Johannes and van Eyk, Erwin and Herbst, Nikolas and Kounev, Samuel},
title = {{Predicting the Costs of Serverless Workflows}},
year = {2020},
isbn = {9781450369916},
publisher = {Association for Computing Machinery},
address = {New York, NY, USA},
url = {https://doi.org/10.1145/3358960.3379133},
doi = {10.1145/3358960.3379133},
booktitle = {Proceedings of the ACM/SPEC International Conference on Performance Engineering},
pages = {265–276},
numpages = {12},
location = {Edmonton AB, Canada},
series = {ICPE '20}
}

@inproceedings{DistServe,
author = {Zhong, Yinmin and Liu, Shengyu and Chen, Junda and Hu, Jianbo and Zhu, Yibo and Liu, Xuanzhe and Jin, Xin and Zhang, Hao},
title = {{DistServe: disaggregating prefill and decoding for goodput-optimized large language model serving}},
year = {2024},
isbn = {978-1-939133-40-3},
publisher = {USENIX Association},
address = {USA},
booktitle = {Proceedings of the 18th USENIX Conference on Operating Systems Design and Implementation},
articleno = {11},
numpages = {18},
location = {Santa Clara, CA, USA},
series = {OSDI'24},
url={https://dl.acm.org/doi/10.5555/3691938.3691949}
}

@inproceedings{cnn_daily_mail_1,
    title = "Get To The Point: Summarization with Pointer-Generator Networks",
    author = "See, Abigail  and
      Liu, Peter J.  and
      Manning, Christopher D.",
    booktitle = "Proceedings of the 55th Annual Meeting of the Association for Computational Linguistics (Volume 1: Long Papers)",
    month = jul,
    year = "2017",
    address = "Vancouver, Canada",
    publisher = "Association for Computational Linguistics",
    url = "https://www.aclweb.org/anthology/P17-1099",
    doi = "10.18653/v1/P17-1099",
    pages = "1073--1083",
}

@inproceedings{cnn_daily_mail_2,
  author={Karl Moritz Hermann and Tomás Kociský and Edward Grefenstette and Lasse Espeholt and Will Kay and Mustafa Suleyman and Phil Blunsom},
  title={{Teaching Machines to Read and Comprehend}},
  year={2015},
  cdate={1420070400000},
  pages={1693-1701},
  url={http://papers.nips.cc/paper/5945-teaching-machines-to-read-and-comprehend},
  booktitle={NIPS},
}

@misc{humaneval,
      title={{Evaluating Large Language Models Trained on Code}}, 
      author={Mark Chen and Jerry Tworek and Heewoo Jun and Qiming Yuan and Henrique Ponde de Oliveira Pinto and Jared Kaplan and Harri Edwards and Yuri Burda and Nicholas Joseph and Greg Brockman and Alex Ray and Raul Puri and Gretchen Krueger and Michael Petrov and Heidy Khlaaf and Girish Sastry and Pamela Mishkin and Brooke Chan and Scott Gray and Nick Ryder and Mikhail Pavlov and Alethea Power and Lukasz Kaiser and Mohammad Bavarian and Clemens Winter and Philippe Tillet and Felipe Petroski Such and Dave Cummings and Matthias Plappert and Fotios Chantzis and Elizabeth Barnes and Ariel Herbert-Voss and William Hebgen Guss and Alex Nichol and Alex Paino and Nikolas Tezak and Jie Tang and Igor Babuschkin and Suchir Balaji and Shantanu Jain and William Saunders and Christopher Hesse and Andrew N. Carr and Jan Leike and Josh Achiam and Vedant Misra and Evan Morikawa and Alec Radford and Matthew Knight and Miles Brundage and Mira Murati and Katie Mayer and Peter Welinder and Bob McGrew and Dario Amodei and Sam McCandlish and Ilya Sutskever and Wojciech Zaremba},
      year={2021},
      eprint={2107.03374},
      archivePrefix={arXiv},
      primaryClass={cs.LG},
      url={https://arxiv.org/abs/2107.03374}, 
}

@inproceedings{agnews_yelp,
author = {Zhang, Xiang and Zhao, Junbo and LeCun, Yann},
title = {{Character-level convolutional networks for text classification}},
year = {2015},
publisher = {MIT Press},
address = {Cambridge, MA, USA},
booktitle = {Proceedings of the 29th International Conference on Neural Information Processing Systems - Volume 1},
pages = {649–657},
numpages = {9},
location = {Montreal, Canada},
series = {NIPS'15},
url={https://dl.acm.org/doi/10.5555/2969239.2969312}
}

@inproceedings{dialogsum,
    title = "{D}ialog{S}um: {A} Real-Life Scenario Dialogue Summarization Dataset",
    author = "Chen, Yulong  and
      Liu, Yang  and
      Chen, Liang  and
      Zhang, Yue",
    booktitle = "Findings of the Association for Computational Linguistics: ACL-IJCNLP 2021",
    month = aug,
    year = "2021",
    address = "Online",
    publisher = "Association for Computational Linguistics",
    url = "https://aclanthology.org/2021.findings-acl.449",
    doi = "10.18653/v1/2021.findings-acl.449",
    pages = "5062--5074",
}

@inproceedings{writingprompts,
    title = "Hierarchical Neural Story Generation",
    author = "Fan, Angela  and
      Lewis, Mike  and
      Dauphin, Yann",
    editor = "Gurevych, Iryna  and
      Miyao, Yusuke",
    booktitle = "Proceedings of the 56th Annual Meeting of the Association for Computational Linguistics (Volume 1: Long Papers)",
    month = jul,
    year = "2018",
    address = "Melbourne, Australia",
    publisher = "Association for Computational Linguistics",
    url = "https://aclanthology.org/P18-1082/",
    doi = "10.18653/v1/P18-1082",
    pages = "889--898"
}

@misc{optuna_tpe_sampler,
  title        = {{TPE Sampler}},
  author       = {{Optuna Team}},
  year         = {2024},
  howpublished = {\url{https://hub.optuna.org/samplers/tpe/}},
}

@article{llama-3,
  title={{The llama 3 herd of models}},
  author={Grattafiori, Aaron and Dubey, Abhimanyu and Jauhri, Abhinav and Pandey, Abhinav and Kadian, Abhishek and Al-Dahle, Ahmad and Letman, Aiesha and Mathur, Akhil and Schelten, Alan and Vaughan, Alex and others},
  journal={arXiv preprint arXiv:2407.21783},
  year={2024},
  url={https://arxiv.org/abs/2407.21783}
}

@misc{qwen-2.5,
      title={{Qwen2.5 Technical Report}}, 
      author={Qwen and : and An Yang and Baosong Yang and Beichen Zhang and Binyuan Hui and Bo Zheng and Bowen Yu and Chengyuan Li and Dayiheng Liu and Fei Huang and Haoran Wei and Huan Lin and Jian Yang and Jianhong Tu and Jianwei Zhang and Jianxin Yang and Jiaxi Yang and Jingren Zhou and Junyang Lin and Kai Dang and Keming Lu and Keqin Bao and Kexin Yang and Le Yu and Mei Li and Mingfeng Xue and Pei Zhang and Qin Zhu and Rui Men and Runji Lin and Tianhao Li and Tianyi Tang and Tingyu Xia and Xingzhang Ren and Xuancheng Ren and Yang Fan and Yang Su and Yichang Zhang and Yu Wan and Yuqiong Liu and Zeyu Cui and Zhenru Zhang and Zihan Qiu},
      year={2025},
      eprint={2412.15115},
      archivePrefix={arXiv},
      primaryClass={cs.CL},
      url={https://arxiv.org/abs/2412.15115}, 
}

@misc{mistral,
      title={{Mistral 7B}}, 
      author={Albert Q. Jiang and Alexandre Sablayrolles and Arthur Mensch and Chris Bamford and Devendra Singh Chaplot and Diego de las Casas and Florian Bressand and Gianna Lengyel and Guillaume Lample and Lucile Saulnier and Lélio Renard Lavaud and Marie-Anne Lachaux and Pierre Stock and Teven Le Scao and Thibaut Lavril and Thomas Wang and Timothée Lacroix and William El Sayed},
      year={2023},
      eprint={2310.06825},
      archivePrefix={arXiv},
      primaryClass={cs.CL},
      url={https://arxiv.org/abs/2310.06825}, 
}

@inproceedings{ALAP,
author = {Sinha, Prasoon and Kaffes, Kostis and Yadwadkar, Neeraja J.},
title = {{ALAP: Intent-Based Serverless Computing via Delayed Decision-Making}},
year = {2026},
isbn = {9798400722769},
publisher = {Association for Computing Machinery},
address = {New York, NY, USA},
url = {https://doi.org/10.1145/3772052.3772262},
doi = {10.1145/3772052.3772262},
booktitle = {Proceedings of the 2025 ACM Symposium on Cloud Computing},
pages = {831–846},
numpages = {16},
location = {
},
series = {SoCC '25}
}

@inproceedings{AQUATOPE,
author = {Zhou, Zhuangzhuang and Zhang, Yanqi and Delimitrou, Christina},
title = {{AQUATOPE: QoS-and-Uncertainty-Aware Resource Management for Multi-stage Serverless Workflows}},
year = {2022},
isbn = {9781450399159},
publisher = {Association for Computing Machinery},
address = {New York, NY, USA},
url = {https://doi.org/10.1145/3567955.3567960},
doi = {10.1145/3567955.3567960},
booktitle = {Proceedings of the 28th ACM International Conference on Architectural Support for Programming Languages and Operating Systems, Volume 1},
pages = {1–14},
numpages = {14},
location = {Vancouver, BC, Canada},
series = {ASPLOS 2023}
}

@inproceedings{Bilal-Serverless,
author = {Bilal, Muhammad and Canini, Marco and Fonseca, Rodrigo and Rodrigues, Rodrigo},
title = {With Great Freedom Comes Great Opportunity: Rethinking Resource Allocation for Serverless Functions},
year = {2023},
isbn = {9781450394871},
publisher = {Association for Computing Machinery},
address = {New York, NY, USA},
url = {https://doi.org/10.1145/3552326.3567506},
doi = {10.1145/3552326.3567506},
booktitle = {Proceedings of the Eighteenth European Conference on Computer Systems},
pages = {381–397},
numpages = {17},
location = {Rome, Italy},
series = {EuroSys '23}
}

@inproceedings{Cypress,
author = {Bhasi, Vivek M. and Gunasekaran, Jashwant Raj and Sharma, Aakash and Kandemir, Mahmut Taylan and Das, Chita},
title = {{Cypress: input size-sensitive container provisioning and request scheduling for serverless platforms}},
year = {2022},
isbn = {9781450394147},
publisher = {Association for Computing Machinery},
address = {New York, NY, USA},
url = {https://doi.org/10.1145/3542929.3563464},
doi = {10.1145/3542929.3563464},
booktitle = {Proceedings of the 13th Symposium on Cloud Computing},
pages = {257–272},
numpages = {16},
location = {San Francisco, California},
series = {SoCC '22}
}

@article{resource_availability_1,
author = {Albalawi, Nasser S.},
title = {{Dynamic scheduling strategies for cloud-based load balancing in parallel and distributed systems}},
year = {2025},
issue_date = {Dec 2025},
publisher = {Hindawi Limited},
address = {London, GBR},
volume = {14},
number = {1},
issn = {2192-113X},
url = {https://doi.org/10.1186/s13677-025-00757-6},
doi = {10.1186/s13677-025-00757-6},
journal = {J. Cloud Comput.},
month = jul,
numpages = {25}
}

@inproceedings{ResourceCentral,
author = {Cortez, Eli and Bonde, Anand and Muzio, Alexandre and Russinovich, Mark and Fontoura, Marcus and Bianchini, Ricardo},
title = {{Resource Central: Understanding and Predicting Workloads for Improved Resource Management in Large Cloud Platforms}},
year = {2017},
isbn = {9781450350853},
publisher = {Association for Computing Machinery},
address = {New York, NY, USA},
url = {https://doi.org/10.1145/3132747.3132772},
doi = {10.1145/3132747.3132772},
booktitle = {Proceedings of the 26th Symposium on Operating Systems Principles},
pages = {153–167},
numpages = {15},
location = {Shanghai, China},
series = {SOSP '17}
}

@inproceedings{planning_ability_1,
title={On the Planning Abilities of Large Language Models - A Critical Investigation},
author={Karthik Valmeekam and Matthew Marquez and Sarath Sreedharan and Subbarao Kambhampati},
booktitle={Thirty-seventh Conference on Neural Information Processing Systems},
year={2023},
url={https://openreview.net/forum?id=X6dEqXIsEW}
}

@article{planning_ability_2,
title={{A Systematic Evaluation of the Planning and Scheduling Abilities of the Reasoning Model o1}},
author={Karthik Valmeekam and Kaya Stechly and Atharva Gundawar and Subbarao Kambhampati},
journal={Transactions on Machine Learning Research},
issn={2835-8856},
year={2025},
url={https://openreview.net/forum?id=FkKBxp0FhR},
note={}
}

@inproceedings{planning_ability_3,
title={{Position: {LLM}s Can{\textquoteright}t Plan, But Can Help Planning in {LLM}-Modulo Frameworks}},
author={Subbarao Kambhampati and Karthik Valmeekam and Lin Guan and Mudit Verma and Kaya Stechly and Siddhant Bhambri and Lucas Paul Saldyt and Anil B Murthy},
booktitle={Forty-first International Conference on Machine Learning},
year={2024},
url={https://openreview.net/forum?id=Th8JPEmH4z}
}

@inproceedings{planning_dark,
author = {Huang, Sukai and Lipovetzky, Nir and Cohn, Trevor},
title = {{Planning in the dark: LLM-symbolic planning pipeline without experts}},
year = {2025},
isbn = {978-1-57735-897-8},
publisher = {AAAI Press},
url = {https://doi.org/10.1609/aaai.v39i25.34855},
doi = {10.1609/aaai.v39i25.34855},
booktitle = {Proceedings of the Thirty-Ninth AAAI Conference on Artificial Intelligence and Thirty-Seventh Conference on Innovative Applications of Artificial Intelligence and Fifteenth Symposium on Educational Advances in Artificial Intelligence},
articleno = {2957},
numpages = {9},
series = {AAAI'25/IAAI'25/EAAI'25}
}

@book{BO_textbook, 
  place={Cambridge}, 
  title={{Bayesian Optimization}}, 
  publisher={Cambridge University Press}, 
  author={Garnett, Roman}, 
  year={2023}
}

@inproceedings{ServerlessLLM,
author = {Fu, Yao and Xue, Leyang and Huang, Yeqi and Brabete, Andrei-Octavian and Ustiugov, Dmitrii and Patel, Yuvraj and Mai, Luo},
title = {ServerlessLLM: low-latency serverless inference for large language models},
year = {2024},
isbn = {978-1-939133-40-3},
publisher = {USENIX Association},
address = {USA},
booktitle = {Proceedings of the 18th USENIX Conference on Operating Systems Design and Implementation},
articleno = {8},
numpages = {19},
location = {Santa Clara, CA, USA},
series = {OSDI'24},
url={https://dl.acm.org/doi/abs/10.5555/3691938.3691946}
}

@InProceedings{BO_org,
author="Mo{\v{c}}kus, J.",
editor="Marchuk, G. I.",
title="On bayesian methods for seeking the extremum",
booktitle="Optimization Techniques IFIP Technical Conference Novosibirsk, July 1--7, 1974",
year="1975",
publisher="Springer Berlin Heidelberg",
address="Berlin, Heidelberg",
pages="400--404",
isbn="978-3-540-37497-8",
url={https://doi.org/10.1007/3-540-07165-2_55}
}

@misc{InferX,
  title        = {{InferX}},
  author       = {{InferX AI}},
  year         = {2026},
  howpublished = {\url{https://inferx.net/}},
}

@book{evolutionary_textbook,
  title={{Introduction to evolutionary computing}},
  author={Eiben, Agoston E and Smith, James E},
  year={2015},
  publisher={Springer},
  url={https://doi.org/10.1007/978-3-662-05094-1}
}

@inproceedings{evolutionary_vis,
author = {Walter, Mathew J. and Walker, David J. and Craven, Matthew J.},
title = {{An explainable visualisation of the evolutionary search process}},
year = {2022},
isbn = {9781450392686},
publisher = {Association for Computing Machinery},
address = {New York, NY, USA},
url = {https://doi.org/10.1145/3520304.3533984},
doi = {10.1145/3520304.3533984},
booktitle = {Proceedings of the Genetic and Evolutionary Computation Conference Companion},
pages = {1794–1802},
numpages = {9},
location = {Boston, Massachusetts},
series = {GECCO '22}
}

@inproceedings{Steiner_1,
author = {Agrawal, Ajit and Klein, Philip and Ravi, R.},
title = {{When trees collide: an approximation algorithm for the generalized Steiner problem on networks}},
year = {1991},
isbn = {0897913973},
publisher = {Association for Computing Machinery},
address = {New York, NY, USA},
url = {https://doi.org/10.1145/103418.103437},
doi = {10.1145/103418.103437},
booktitle = {Proceedings of the Twenty-Third Annual ACM Symposium on Theory of Computing},
pages = {134–144},
numpages = {11},
location = {New Orleans, Louisiana, USA},
series = {STOC '91}
}

@inproceedings{Steiner_2,
author = {Goemans, Michel X. and Williamson, David P.},
title = {{A general approximation technique for constrained forest problems}},
year = {1992},
isbn = {089791466X},
publisher = {Society for Industrial and Applied Mathematics},
address = {USA},
booktitle = {Proceedings of the Third Annual ACM-SIAM Symposium on Discrete Algorithms},
pages = {307–316},
numpages = {10},
location = {Orlando, Florida, USA},
series = {SODA '92},
url={https://dl.acm.org/doi/10.5555/139404.139468}
}

@book{BFS,
  title={The Shortest Path Through a Maze},
  author={Moore, E.F.},
  series={Bell Telephone System. Technical publications. monograph},
  url={https://books.google.co.in/books?id=IVZBHAAACAAJ},
  year={1959},
  publisher={Bell Telephone System.}
}

@book{graph_search_textbook,
  title={Introduction to algorithms},
  author={Cormen, Thomas H and Leiserson, Charles E and Rivest, Ronald L and Stein, Clifford},
  year={2022},
  publisher={MIT press},
  url={https://mitpress.mit.edu/9780262367509/introduction-to-algorithms/}
}

@article{DFS,
author = {Tarjan, Robert},
title = {Depth-First Search and Linear Graph Algorithms},
journal = {SIAM Journal on Computing},
volume = {1},
number = {2},
pages = {146-160},
year = {1972},
doi = {10.1137/0201010},
URL = {https://doi.org/10.1137/0201010},
eprint = {https://doi.org/10.1137/0201010}
}

@inbook{dijkstra,
author = {Dijkstra, E. W.},
title = {A Note on Two Problems in Connexion with Graphs},
year = {2022},
isbn = {9781450397735},
publisher = {Association for Computing Machinery},
address = {New York, NY, USA},
edition = {1},
url = {https://doi.org/10.1145/3544585.3544600},
booktitle = {Edsger Wybe Dijkstra: His Life,Work, and Legacy},
pages = {287–290},
numpages = {4}
}

@ARTICLE{Astar,
  author={Hart, Peter E. and Nilsson, Nils J. and Raphael, Bertram},
  journal={IEEE Transactions on Systems Science and Cybernetics}, 
  title={{A Formal Basis for the Heuristic Determination of Minimum Cost Paths}}, 
  year={1968},
  volume={4},
  number={2},
  pages={100-107},
  doi={10.1109/TSSC.1968.300136},
  url={https://doi.org/10.1109/TSSC.1968.300136}}

@book{Astar_textbook,
author = {Russell, Stuart and Norvig, Peter},
title = {Artificial Intelligence: A Modern Approach},
year = {2009},
isbn = {0136042597},
publisher = {Prentice Hall Press},
address = {USA},
edition = {3rd},
url={https://dl.acm.org/doi/10.5555/1671238}
}

@inproceedings{online-tree-deletions,
author = {Gupta, Anupam and Kumar, Amit},
title = {{Online steiner tree with deletions}},
year = {2014},
isbn = {9781611973389},
publisher = {Society for Industrial and Applied Mathematics},
address = {USA},
booktitle = {Proceedings of the Twenty-Fifth Annual ACM-SIAM Symposium on Discrete Algorithms},
pages = {455–467},
numpages = {13},
location = {Portland, Oregon},
series = {SODA '14},
url={https://dl.acm.org/doi/10.5555/2634074.2634108}
}

@misc{BO_limitation_1,
      title={{A Tutorial on Bayesian Optimization}}, 
      author={Peter I. Frazier},
      year={2018},
      eprint={1807.02811},
      archivePrefix={arXiv},
      primaryClass={stat.ML},
      url={https://arxiv.org/abs/1807.02811}, 
}

@inproceedings{BO_limitation_2,
title={Understanding High-Dimensional Bayesian Optimization},
author={Leonard Papenmeier and Matthias Poloczek and Luigi Nardi},
booktitle={Forty-second International Conference on Machine Learning},
year={2025},
url={https://openreview.net/forum?id=1d2fpvyKvJ}
}

@inproceedings{BO_limitation_3,
  title={Bayesian optimization of combinatorial structures},
  author={Baptista, Ricardo and Poloczek, Matthias},
  booktitle={International conference on machine learning},
  pages={462--471},
  year={2018},
  organization={PMLR},
  url={https://proceedings.mlr.press/v80/baptista18a/baptista18a.pdf}
}

@misc{genAI_app_1,
  author       = {{Data and Beyond}},
  title        = {Practical Generative AI: 9 NLP Tasks with Hugging Face Transformers},
  year         = {2025},
  howpublished = {\url{https://medium.com/data-and-beyond/practical-generative-ai-9-nlp-tasks-with-hugging-face-transformers-2ec7754a76e2}},
  note         = {Medium article, accessed April 2026}
}

@article{genAI_app_2,
  title={{Natural language processing: state of the art, current trends and challenges}},
  author={Khurana, Diksha and Koli, Aditya and Khatter, Kiran and Singh, Sukhdev},
  journal={Multimedia tools and applications},
  volume={82},
  number={3},
  pages={3713--3744},
  year={2023},
  publisher={Springer},
  url={https://doi.org/10.1007/s11042-022-13428-4}
}

@misc{genAI_app_3,
  author       = {{AIMonk Team}},
  title        = {{10+ Amazing Natural Language Processing Examples in 2026}},
  year         = {2026},
  howpublished = {\url{https://aimonk.com/natural-language-processing-examples/}},
  note         = {Published March 10, 2026; accessed April 2026}
}

@inproceedings{LLMRouting,
title={{Universal Model Routing for Efficient {LLM} Inference}},
author={Wittawat Jitkrittum and Harikrishna Narasimhan and Ankit Singh Rawat and Jeevesh Juneja and Congchao Wang and Zifeng Wang and Alec Go and Chen-Yu Lee and Pradeep Shenoy and Rina Panigrahy and Aditya Krishna Menon and Sanjiv Kumar},
booktitle={The Fourteenth International Conference on Learning Representations},
year={2026},
url={https://openreview.net/forum?id=ka82fvJ5f1}
}

@inproceedings {Orion,
author = {Ashraf Mahgoub and Edgardo Barsallo Yi and Karthick Shankar and Sameh Elnikety and Somali Chaterji and Saurabh Bagchi},
title = {{ORION} and the Three Rights: Sizing, Bundling, and Prewarming for Serverless {DAGs}},
booktitle = {16th USENIX Symposium on Operating Systems Design and Implementation (OSDI 22)},
year = {2022},
isbn = {978-1-939133-28-1},
address = {Carlsbad, CA},
pages = {303--320},
url = {https://www.usenix.org/conference/osdi22/presentation/mahgoub},
publisher = {USENIX Association},
month = jul
}

@inproceedings{FaaSFlow,
author = {Li, Zijun and Liu, Yushi and Guo, Linsong and Chen, Quan and Cheng, Jiagan and Zheng, Wenli and Guo, Minyi},
title = {{FaaSFlow: enable efficient workflow execution for function-as-a-service}},
year = {2022},
isbn = {9781450392051},
publisher = {Association for Computing Machinery},
address = {New York, NY, USA},
url = {https://doi.org/10.1145/3503222.3507717},
doi = {10.1145/3503222.3507717},
booktitle = {Proceedings of the 27th ACM International Conference on Architectural Support for Programming Languages and Operating Systems},
pages = {782–796},
numpages = {15},
location = {Lausanne, Switzerland},
series = {ASPLOS '22}
}

@inproceedings{Catalyzer,
author = {Du, Dong and Yu, Tianyi and Xia, Yubin and Zang, Binyu and Yan, Guanglu and Qin, Chenggang and Wu, Qixuan and Chen, Haibo},
title = {{Catalyzer: Sub-millisecond Startup for Serverless Computing with Initialization-less Booting}},
year = {2020},
isbn = {9781450371025},
publisher = {Association for Computing Machinery},
address = {New York, NY, USA},
url = {https://doi.org/10.1145/3373376.3378512},
doi = {10.1145/3373376.3378512},
booktitle = {Proceedings of the Twenty-Fifth International Conference on Architectural Support for Programming Languages and Operating Systems},
pages = {467–481},
numpages = {15},
location = {Lausanne, Switzerland},
series = {ASPLOS '20}
}

@inproceedings{REAP,
author = {Ustiugov, Dmitrii and Petrov, Plamen and Kogias, Marios and Bugnion, Edouard and Grot, Boris},
title = {{Benchmarking, analysis, and optimization of serverless function snapshots}},
year = {2021},
isbn = {9781450383172},
publisher = {Association for Computing Machinery},
address = {New York, NY, USA},
url = {https://doi.org/10.1145/3445814.3446714},
doi = {10.1145/3445814.3446714},
booktitle = {Proceedings of the 26th ACM International Conference on Architectural Support for Programming Languages and Operating Systems},
pages = {559–572},
numpages = {14},
location = {Virtual, USA},
series = {ASPLOS '21}
}

@inproceedings{FaaSnap,
author = {Ao, Lixiang and Porter, George and Voelker, Geoffrey M.},
title = {{FaaSnap: FaaS made fast using snapshot-based VMs}},
year = {2022},
isbn = {9781450391627},
publisher = {Association for Computing Machinery},
address = {New York, NY, USA},
url = {https://doi.org/10.1145/3492321.3524270},
doi = {10.1145/3492321.3524270},
booktitle = {Proceedings of the Seventeenth European Conference on Computer Systems},
pages = {730–746},
numpages = {17},
location = {Rennes, France},
series = {EuroSys '22}
}

@inproceedings {ServerlessWild,
author = {Mohammad Shahrad and Rodrigo Fonseca and Inigo Goiri and Gohar Chaudhry and Paul Batum and Jason Cooke and Eduardo Laureano and Colby Tresness and Mark Russinovich and Ricardo Bianchini},
title = {{Serverless in the Wild: Characterizing and Optimizing the Serverless Workload at a Large Cloud Provider}},
booktitle = {2020 USENIX Annual Technical Conference (USENIX ATC 20)},
year = {2020},
isbn = {978-1-939133-14-4},
pages = {205--218},
url = {https://www.usenix.org/conference/atc20/presentation/shahrad},
publisher = {USENIX Association},
month = jul
}
\bibliographystyle{iclr2027_conference}

\newpage
\appendix
\section{Appendix Organization}
The appendix is organized as follows.
Appendix~\ref{app:related_works} extends the related work, positioning \texttt{FlowMaker} relative to workflow construction, model routing, generative-AI workflow serving, and serverless workflow systems.
Appendix~\ref{app:model_characterization} characterizes the three dimensions of the workflow-variant space: workflows, models, and deployment configurations.
Appendix~\ref{sec:appendix_alog_alternatives} discusses alternative selection approaches (lookup tables, Bayesian optimization and evolutionary search, and graph algorithms) and their limitations.
Appendix~\ref{sec:appendix_algo} summarizes notation, describes progressive SLO relaxation, and analyzes the complexity of Algorithm~\ref{algo:wvs}.
Appendix~\ref{app:implementation_details} details \texttt{FlowMaker}'s APIs and prototype implementation.
Appendix~\ref{app:experimental_details} describes the experimental setup, including trace construction, tasks, the model catalog, baseline configurations, SLO definitions, the testbed, and metrics.
Appendix~\ref{app:extra_eval} reports additional results: a per-baseline breakdown of the end-to-end comparison (\S\ref{app:e2e_detailed}), scaling under fixed cluster capacity (\S\ref{sec:eval_scalability}), sensitivity to latency SLOs (\S\ref{sec:slo_sensitivity}), and model-loading overheads (\S\ref{sec:overheads}).
\section{Extended Related Work}
\label{app:related_works}

\noindent\textbf{Workflow construction and application-level optimization.}
Prior work automates different aspects of generative-AI application design. DSPy~\citep{DSPy} compiles declarative LM programs by optimizing prompts and demonstrations. AutoFlow~\citep{AutoFlow} generates natural-language agent workflows using fine-tuning- and in-context-based optimization, while ADAS~\citep{ADAS} searches over code-defined agentic systems and AFlow~\citep{AFlow} refines code-represented workflows using Monte Carlo tree search and execution feedback. Cognify~\citep{Cognify} autotunes an existing workflow through hierarchical search over structural changes, step operators---including model choice---and prompts, using workflow-level measurements of output quality, latency, and monetary cost. DAAO~\citep{DAAO} instead adapts workflow depth, reasoning operators, and heterogeneous model assignments to individual queries based on their estimated difficulty.

These systems optimize application logic, prompts, or per-query model use. They do not jointly choose physical deployment configurations, such as tensor- and pipeline-parallelism, or determine whether stages should use new or already deployed model instances based on current load and GPU availability. They are therefore complementary to {\system}: their generated or tuned workflows can supply candidate workflows, after which {\system} selects a deployable {\wfvar} and reuse plan at task-deployment time.

\noindent\textbf{Model selection and routing.}
LLMSelector~\citep{LLMSelector} assigns models to modules in a fixed compound-AI architecture. RouteLLM~\citep{RouteLLM} routes each query between a fixed strong--weak model pair, while MasRouter~\citep{MasRouter} additionally selects collaboration modes, agent roles, and models for multi-agent systems. Other LLM routers similarly select among available models to balance output quality and inference cost~\citep{LLMRouting}. These approaches determine which model processes an input, but do not jointly select alternative workflow decompositions, physical deployment configurations, and new-versus-reused model instances under current cluster state. They can be combined with {\system} to route individual requests among the instances selected when a task is deployed.

\noindent\textbf{Generative-AI workflow serving.}
Aragog~\citep{Aragog} assumes a supplied workflow DAG and a configured pool of model-serving engines. For each request, it identifies accuracy-preserving model assignments and selects among them stage by stage using current engine load. Thus, Aragog adapts model routing during the execution of a fixed workflow but does not select the workflow structure or determine which model configurations and instances should be deployed.

Murakkab~\citep{Murakkab} exposes a declarative workflow interface and uses offline workflow and model profiles with periodic MILP optimization. Its optimizer selects workflow parameters, executors, model and hardware configurations, parallelism strategies, and instance assignments under predicted load. Murakkab can multiplex model instances across workflows and periodically reconfigure its deployment as demand and resource availability change. {\system} addresses a different decision scope and timescale: when a task is deployed, it selects among user-approved alternative workflows and their induced model and deployment choices, and maps each stage to a new or compatible deployed instance using the observed cluster state. Subsequent requests use this selected deployment. Aragog's request-time routing and Murakkab's periodic resource planning are therefore complementary to {\system}'s task-deployment-time workflow-variant and reuse selection.

\noindent\textbf{Workflow orchestration and agent execution.}
Application frameworks such as LangChain, LangGraph, CrewAI, AutoGen, and MetaGPT provide abstractions for composing LLM calls, tools, agents, state, and control flow~\citep{LangChain, LangGraph, CrewAI, AutoGen, MetaGPT}. Their programming models range from explicitly structured workflows to conversational multi-agent coordination, but the application structure or agent logic is supplied by the developer.

Systems work provides complementary execution-time mechanisms. Parrot exposes dependencies among LLM calls through semantic variables to enable application-aware scheduling and context sharing~\citep{Parrot}. Nalar uses futures and managed state to support policy-driven scheduling, routing, and resource management for dynamic agent programs~\citep{Nalar}, while AIOS provides services for agent scheduling, context, memory, storage, and tool management~\citep{AIOS}. KVFlow and Helium exploit workflow structure to improve cache management, prefetching, or scheduling~\citep{KVFlow, Helium}. These systems improve the construction or execution of an application, whereas {\system} selects its deployable workflow variant and reuse plan under task-level SLOs and current cluster conditions.

\noindent\textbf{LLM inference, profiling, and deployment.}
Inference runtimes such as vLLM, TensorRT-LLM, and DeepSpeed provide complementary mechanisms for efficiently executing a selected model, including KV-cache management, batching, optimized kernels, and distributed parallelism~\citep{vllm, TensorRT-LLM, deepspeed}. Profiling and performance-modeling systems reduce the cost of evaluating deployment choices. MaverIQ estimates latency and memory across parallelism configurations through lightweight profiling~\citep{MaverIQ}; Vidur simulates LLM inference and supports cost-effective deployment selection~\citep{Vidur}; NeuSight forecasts model latency across model--GPU combinations~\citep{NeuSight}; and MAD-Max models distributed mappings and parallelization strategies~\citep{MAD_Max}. AlpaServe jointly selects placement and model parallelism for a collection of models under bursty demand~\citep{AlpaServe}.

{\system} builds on these execution and prediction mechanisms but operates at a different optimization scope. Rather than optimizing the deployment of a predetermined model or model collection, it selects an end-to-end workflow variant, including workflow structure and accuracy-dependent model choices, and accounts for reuse across tasks.

\noindent\textbf{Serverless workflow resource management.}
Serverless workflow systems execute user-specified DAGs of functions under latency and cost objectives. Aquatope~\citep{AQUATOPE} and Cypress~\citep{Cypress} provision workflow stages under performance uncertainty; Orion~\citep{Orion} sizes, bundles, and prewarms stages; FaaSFlow~\citep{FaaSFlow} optimizes distributed workflow execution and data movement; and ALAP~\citep{ALAP} delays resource-allocation decisions to improve efficiency. These systems generally treat the workflow structure and function implementations as fixed and optimize how their resources are provisioned, placed, or scheduled.

Generative-AI workflow serving introduces additional decisions. Different models provide different output quality, each model admits multiple parallel configurations, and reusing an instance changes its load and capacity available to other tasks. LLM instances also have large, persistent GPU-memory footprints and may span multiple GPUs. Finally, the serving system must select among alternative user-approved workflows rather than receiving a single fixed DAG. Serverless resource-management mechanisms remain complementary, but their fixed-workflow formulations do not directly solve this joint selection problem. We nevertheless compare against the Bayesian-optimization search strategy used by serverless systems~\citep{AQUATOPE, Bilal-Serverless} in \S~\ref{sec:eval_algo-efficiency}.

\noindent\textbf{Cold starts and model loading.}
A complementary line of work reduces the time required to initialize serverless functions or load model state. Prior systems mitigate cold starts and accelerate restoration~\citep{Catalyzer, REAP, FaaSnap}, prewarm instances~\citep{Orion, ServerlessWild}, or optimize checkpoint loading and live migration for LLM serving~\citep{ServerlessLLM}. {\system} addresses the preceding decision: it determines which models and configurations to deploy and whether an already deployed instance can be reused without exceeding its capacity. These techniques can accelerate any new deployment selected by {\system}.

\section{Workflow-Variant Space Characterization}
\label{app:model_characterization}

We provide additional characterization of the three dimensions forming the workflow-variant space.

\begin{figure}[h]
    \centering
    \includegraphics[width=.65\linewidth]{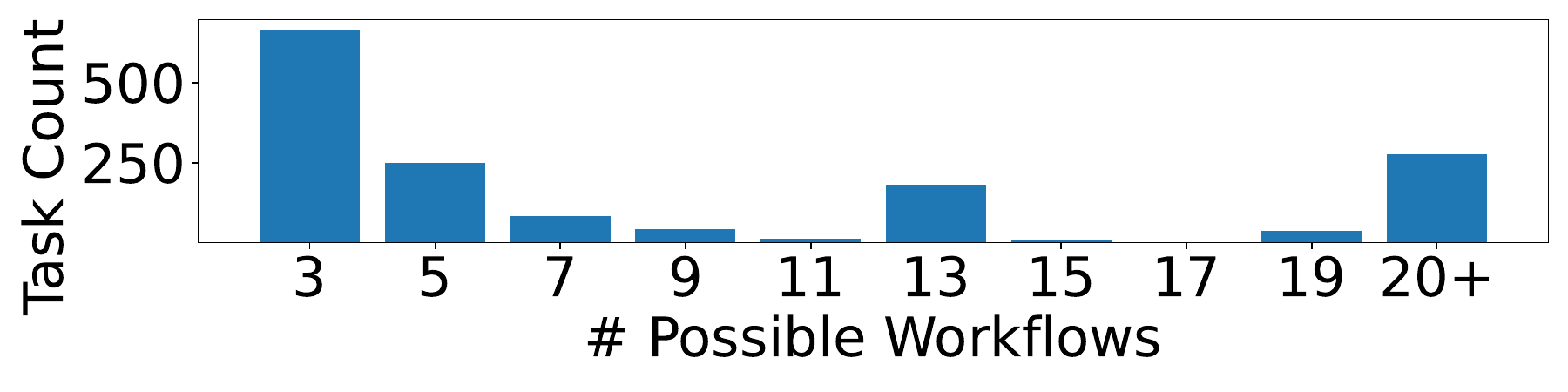}
    \vspace{-3mm}
    \caption{Each task in the AsyncHow dataset can be executed using multiple workflows, each with distinct characteristics. Those choices expand the search space.}
    \label{fig:possible_workflows}
\end{figure}

\noindent\textbf{Workflows.}
A task can admit multiple candidate workflows that differ in their stages, dependencies, or execution order. Our analysis of approximately 1.6K tasks in AsyncHow~\citep{AsyncHow} finds that 42\% have at most three candidate workflows, 18\% have at least 20, and the median is five~\citep{Aragog, SWE-Agent} (see Figure~\ref{fig:possible_workflows}). This diversity expands the selection space but also creates opportunities to choose workflows better suited to the task's SLOs and the current cluster state.

\begin{figure}[h]
\centering
\includegraphics[width=0.7\linewidth]{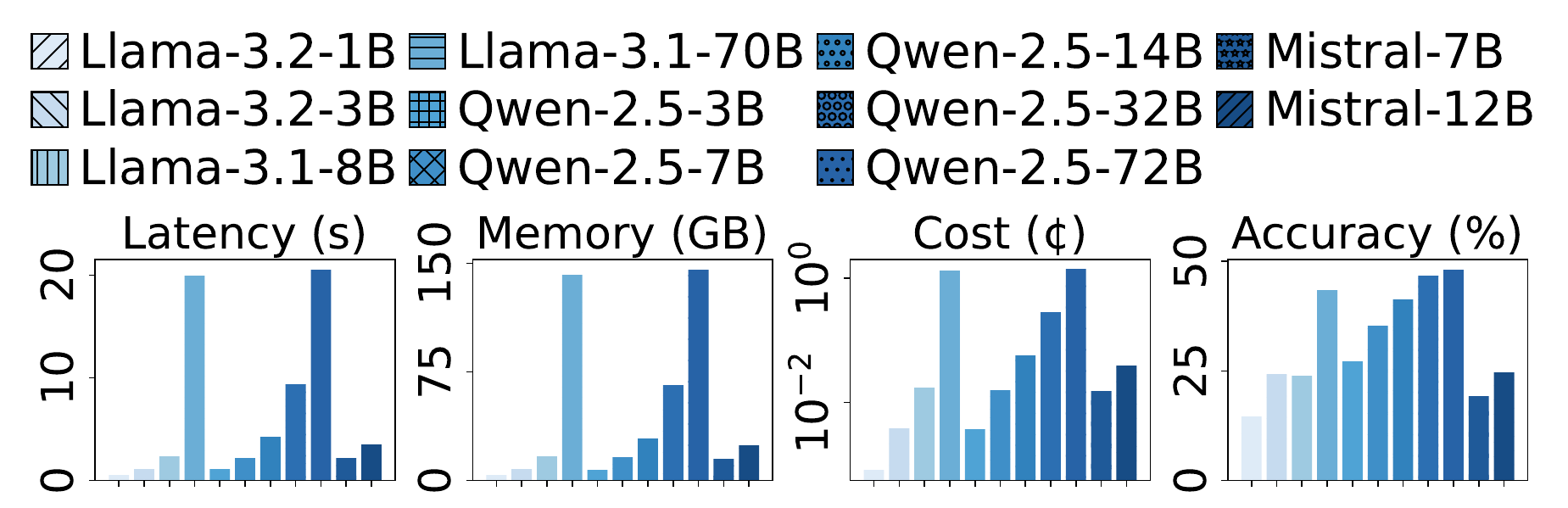}
\caption{
Latency, GPU-memory consumption, serving cost, and benchmark-derived proxy accuracy of the 11 evaluated models. Latency, memory, and cost are measured while serving a query with an output length of 100 tokens.
}
\label{fig:model_selection_effect}
\end{figure}

\noindent\textbf{Models.}
Each LLM stage can be served by models with different latency, GPU-memory, serving-cost, and proxy-accuracy characteristics. Figure~\ref{fig:model_selection_effect} characterizes the 11 models used in our evaluation. Latency, memory, and cost are measured while serving a query with an output length of 100 tokens. Proxy accuracy is the model's average score across six Open LLM Leaderboard benchmarks~\citep{OpenLLM_leaderboard}, rather than a direct evaluation of generated outputs. The models occupy different points across these metrics, requiring selection to account for both user SLOs and resource requirements.

\noindent\textbf{Deployment configurations.}
Even after selecting a model, its tensor- and pipeline-parallelism configuration can substantially change latency, memory requirements, capacity, and cost~\citep{MaverIQ}. On our eight-GPU testbed, Llama-3.1-8B supports 10 evaluated configurations, with up to $3.5\times$ variation in latency and $3.6\times$ variation in cost. Thus, selecting a model without considering its deployment configuration can eliminate otherwise feasible or less expensive workflow variants.

\section{Design Alternatives and their Limitations}
\label{sec:appendix_alog_alternatives}

\noindent\textbf{Lookup table.}
A straightforward approach constructs a lookup table per task that records latency, throughput, and memory for all workflow variants, selecting the one that meets the objective. However, populating this table is expensive, even with state-of-the-art profilers, such as MaverIQ~\citep{MaverIQ}, NeuSight~\citep{NeuSight}, or Vidur~\citep{Vidur}, which only reduce per-model profiling cost. For a simple translate-and-summarize task (3 workflows, 11 models, 10 configurations), this requires evaluating over 24K variants and takes $\sim40$ CPU-minutes with MaverIQ~\citep{MaverIQ}. The cost further increases with reuse, as the space expands to include previously deployed models, making this approach impractical in an online setting.

\noindent\textbf{Traditional search algorithms.}
More sophisticated approaches avoid exhaustive enumeration. Bayesian optimization (BO) and evolutionary search are widely used~\citep{BO_org, BO_textbook, AQUATOPE, Composer}. BO builds a probabilistic surrogate (e.g., Gaussian processes) but incurs modeling overhead and scales poorly to high-dimensional, discrete spaces~\citep{BO_limitation_1, BO_limitation_2, BO_limitation_3}. Evolutionary methods avoid explicit modeling but require many evaluations to converge~\citep{evolutionary_textbook}. Both require careful tuning of surrogates, acquisition functions, and hyperparameters, and incur high decision latency—tens of seconds to minutes—to identify a suitable variant (\S~\ref{sec:eval_algo-efficiency}). This makes them unsuitable for our setting, where selection lies on the critical path.

\noindent\textbf{Graph algorithms.}
Graph-based formulations (e.g., Steiner Forest~\citep{Steiner_1, Steiner_2, online-tree-deletions}) assume a pre-constructed graph, ignoring the cost of building it. In generative AI workflows, the graph $\mathcal{G}$—capturing all workflow-variants—must be constructed on-the-fly, as subtasks and workflows are not known a priori. Even for a single task $\mathcal{T}$, without reusing deployed models, constructing $\mathcal{G}$ costs $O(|\mathcal{W}| \cdot (|\mathcal{M}| \cdot |\mathcal{C}|)^{|\mathcal{S}|})$, where $\mathcal{W}$, $\mathcal{S}$, $\mathcal{M}$, and $\mathcal{C}$ denote workflows, subtasks, models, and configurations. 
This makes runtime graph construction prohibitive and precludes classical traversal algorithms (BFS~\citep{BFS, graph_search_textbook}, DFS~\citep{DFS, graph_search_textbook}, Dijkstra~\citep{dijkstra, graph_search_textbook}, A*~\citep{Astar, Astar_textbook}).
\section{Notation, SLO Relaxation \& Complexity of {\system}'s Algorithm}
\label{sec:appendix_algo}

\noindent\textbf{Notation.}
Table~\ref{tab:notation} summarizes the symbols used in the problem formulation (\S~\ref{subsec:wf_algo}) and in Algorithm~\ref{algo:wvs}. Throughout, a \emph{node} $n_i$ denotes a deployed (model, configuration) instance together with its current serving load, so a {\wfvar} $\mathcal{WV} = \langle \mathbf{n}, \mathbf{gpu} \rangle$ fixes one node per subtask and the GPUs each node occupies.

\begin{table}[h]
\centering
\caption{Notation.}
\label{tab:notation}
\small
\setlength{\tabcolsep}{8pt}
\begin{tabular}{@{}llll@{}}
\toprule
\textbf{Symbol} & \textbf{Description} & \textbf{Symbol} & \textbf{Description} \\
\midrule
$\mathcal{T}$   & Task (in natural language)                    & $\mathbf{n}$   & Nodes $\{n_i\}$ \\
$\mathcal{W}$   & Set of workflows                              & $n_{id}$       & Node (model, config) \\
$\mathcal{S}$   & Set of subtasks                               & $id$           & Node's ID \\
$\mathcal{M}$   & Set of models                                 & $\mathbf{gpu}$ & GPU mapping \\
$\mathcal{C}$   & Set of configs (TP, PP)                       & $\lambda$      & Latency SLO \\
$\mathcal{D}$   & Set of deployed LLMs                          & $\alpha$       & Accuracy SLO \\
$\mathcal{WV}$  & Workflow-variant $(\mathbf{n}, \mathbf{gpu})$ & $\rho$         & Arrival rate \\
\bottomrule
\end{tabular}
\end{table}

\noindent\textbf{Progressive SLO relaxation.}
While users may specify both latency and accuracy SLOs, satisfying both may be infeasible (e.g., no model is both fast and accurate enough). We therefore relax constraints progressively, preserving user intent as much as possible. We first drop the accuracy constraint and select the highest-accuracy workflow variant that meets the latency SLO. If no variant satisfies the latency SLO, we select the lowest-latency variant; if multiple variants have the same latency, we choose the one with the highest accuracy. If only an accuracy SLO is specified and no variant satisfies it, {\system} selects the highest-accuracy variant. If no variant is feasible, {\system} reports that no deployable workflow exists under the given conditions.

\noindent\textbf{Time complexity analysis.}
A brute-force approach enumerates all workflow-variants across all workflows, assigning each subtask a model, deployment configuration, and reuse option. For $|\mathcal{W}|$ workflows, this requires enumerating the Cartesian product across $|\mathcal{S}|$ subtasks, each with $|\mathcal{M}|$ model choices, $|\mathcal{C}|$ configuration options, and $|\mathcal{D}|$ reuse candidates. This yields a complexity of $O\left(|\mathcal{W}| \cdot (|\mathcal{M}| \cdot |\mathcal{C}| \cdot |\mathcal{D}|)^{|\mathcal{S}|}\right)$, which is exponential in $|\mathcal{S}|$.

{\system} makes this tractable by shrinking the base of the exponent through pruning: (i) pruning by accuracy reduces $|\mathcal{M}| \to |\mathcal{M}'|$, (ii) pruning by operational cost reduces $|\mathcal{D}| \to |\mathcal{D}'|$, and (iii) pruning by latency, capacity, and memory constraints reduces $|\mathcal{C}| \to |\mathcal{C}'|$.
We also exit early, avoiding full enumeration of the remaining search space after pruning. Hence, the effective complexity becomes $O\left(|\mathcal{W}| \cdot (|\mathcal{M}'| \cdot |\mathcal{C}'| \cdot |\mathcal{D}'|)^{|\mathcal{S}|}\right)$, where $|\mathcal{M}'| \ll |\mathcal{M}|$, $|\mathcal{C}'| \ll |\mathcal{C}|$, and $|\mathcal{D}'| \ll |\mathcal{D}|$. This significantly reduces the base of the exponent.
In \S~\ref{sec:eval_algo-efficiency}, we empirically demonstrate the time-efficiency of our algorithm.
\section{APIs \& Implementation Details}
\label{app:implementation_details}

\noindent\textbf{APIs.}
Table~\ref{tab:APIs} lists {\system}'s three APIs. Operators call \texttt{register\_model} to add a model to the catalog, optionally supplying a measured accuracy score; if omitted, {\system} falls back to the Open LLM Leaderboard proxy (see Appendix~\ref{app:experimental_details}). Users call \texttt{deploy\_task} with the task in natural language and its latency and accuracy SLOs, after which {\system} generates candidate workflows, selects a {\wfvar}, and deploys it. Expert users may instead pass a \texttt{workflow\_variant} directly, pinning the workflow, models, and configurations; {\system} then honors those constraints while still reusing matching deployed instances. Once a task is deployed, \texttt{inference\_task} submits requests against it, and the Request Orchestrator routes each through the task's selected stages.

\begin{table}[ht!]
\small
\centering 
\caption{{\system}'s APIs.}
\label{tab:APIs}
\begin{tabular}{ll} 
\toprule
API               & Parameters \\ 
\midrule
\texttt{register\_model}   & model, [accuracy]\\
\texttt{deploy\_task}      & task, latSLO, accSLO $\mid$ workflow\_variant \\  
\texttt{inference\_task}   & task, inputText \\  
\bottomrule
\end{tabular}
\end{table}

\noindent\textbf{Prototype architecture.}
We implement {\system} in approximately 3K lines of Python atop TensorRT-LLM~\citep{TensorRT-LLM}. {\system} consists of a logically centralized Controller and a set of GPU workers. The Controller hosts the Workflow Generator, Workflow-Variant Selector, Profiler, Deployer, and Request Orchestrator. Each worker runs a lightweight Monitoring Daemon, and the components communicate through TCP sockets.

\noindent\textbf{Profiling and metadata.}
When a model is registered, the Profiler uses MaverIQ's fingerprint-based profiling~\citep{MaverIQ} to estimate its latency and memory consumption across supported tensor- and pipeline-parallelism configurations. Profiles are approximately 30KB and are stored as CSV files, while model weights are stored in the Model Repository. Profiles are computed once and reused across task deployments.

\noindent\textbf{Workflow generation and selection.}
The Workflow Generator runs as a CPU process. It first queries the Workflow Database, an on-disk JSON catalog of previously approved decompositions. On a cache miss, it prompts OpenAI's \texttt{gpt-oss-120b}~\citep{openai_gpt_oss_120b_api} using chain-of-thought prompting~\citep{CoT} to generate candidate workflows, which are presented to the user for approval. Approved workflows are added to the database.

The Workflow-Variant Selector also runs as a CPU process. It maintains an in-memory view of deployed model instances, their load, and available GPU memory. At task deployment, it combines this state with stored model profiles, user-approved workflows, and the task's SLOs to execute Algorithm~\ref{algo:wvs}.

\noindent\textbf{Deployment and request execution.}
For each model instance that cannot be served through reuse, the Deployer launches a TensorRT-LLM process on the selected GPUs using the chosen parallelism configuration. Each process is assigned an IP/port endpoint stored in an in-memory routing directory. Additional mappings associate tasks with workflow stages and stages with model processes.

The Request Orchestrator serves requests in first-come-first-served order. It looks up the endpoint for the first stage, attaches 100--1000 bytes of routing metadata, and forwards the request. This metadata identifies subsequent stages and their endpoints, allowing intermediate outputs to flow directly between model processes over TCP and keeping the Controller off the inter-stage data path.

\noindent\textbf{Cluster monitoring.}
Each Monitoring Daemon contains two CPU components. The Memory Monitor samples per-GPU memory availability using NVML~\citep{NVML}. The Load Monitor computes each deployed model's load over a two-minute sliding window. Both update the Controller's global in-memory state every five seconds. The selector uses the latest snapshot when evaluating reuse and placement.
\section{Experimental Details}
\label{app:experimental_details}

\noindent\textbf{Trace construction.}
We derive request arrivals from the Azure code and conversation traces~\citep{Splitwise}. We construct four relative offered-load levels by discretizing each trace into intervals, aggregating adjacent intervals, and applying the factors in Table~\ref{tab:trace_scaling}. These percentages denote relative levels within each trace, rather than absolute cluster utilization.

\begin{table}[h]
\centering
\small
\caption{Trace-scaling factors used for each offered-load level.}
\label{tab:trace_scaling}
\begin{tabular}{lcccc}
\toprule
\textbf{Trace} & \textbf{25\%} & \textbf{50\%} & \textbf{75\%} & \textbf{100\%} \\
\midrule
Code         & 0.100 & 0.133 & 0.167 & 0.200 \\
Conversation & 0.050 & 0.067 & 0.083 & 0.100 \\
\bottomrule
\end{tabular}
\end{table}

The resulting workloads contain 212--484 requests. Because the traces provide timestamps and input/output token counts but no semantic inputs, we pair their entries with examples from CNN/DailyMail~\citep{cnn_daily_mail_1, cnn_daily_mail_2}, HumanEval~\citep{humaneval}, Yelp Reviews and AG News~\citep{agnews_yelp}, DialogSum~\citep{dialogsum}, and WritingPrompts~\citep{writingprompts}.

\noindent\textbf{Tasks.}
We evaluate seven representative generative-AI tasks~\citep{genAI_app_1, genAI_app_2, genAI_app_3}. Each task has three candidate workflows containing one to three stages. The tasks span six categories: (1) cross-lingual understanding, which translates and summarizes text across languages; (2) text processing, which summarizes and simplifies text; (3) analysis and reporting, which extracts key phrases and sentiment to synthesize a report; (4) code documentation; (5) opinion mining over product reviews; and (6) document classification and tagging.

\noindent\textbf{Model catalog.}
Our catalog contains 11 open-weight models from three families: Llama-3.2-{1B, 3B} and Llama-3.1-{8B, 70B}~\citep{llama-3}; Qwen2.5-{3B, 7B, 14B, 32B, 72B}~\citep{qwen-2.5}; and Mistral-{7B, 12B}~\citep{mistral}. We treat Llama-3.1-70B and Qwen2.5-72B as fallback models and the remaining nine as the primary catalog. We restrict Llama-3.2-1B/3B and Qwen2.5-3B to stages that perform a single operation because these models do not reliably execute compound operations such as joint translation and summarization.

We construct each stage's candidate-model set using a score that combines model quality from the Open LLM Leaderboard~\citep{OpenLLM_leaderboard}, a logarithmic parameter-count penalty favoring deployment efficiency, and a model-family diversity bonus. This procedure provides multiple candidate models per stage when possible.

\noindent\textbf{LangChain.}
LangChain~\citep{LangChain} represents static workflow execution. We port its execution logic to TensorRT-LLM and provide each task's default workflow, the smallest model satisfying its proxy-accuracy SLO, and TP${}=1$, PP${}=1$. These choices remain fixed as cluster conditions change.

\noindent\textbf{RouteLLM.}
RouteLLM~\citep{RouteLLM} performs request-level routing over a fixed workflow. For every task, we pre-deploy Qwen-2.5-14B and Qwen-2.5-7B as its strong/weak pair and use RouteLLM's pretrained BERT router. A request uses the routed model for every LLM stage in its workflow. We evaluate three routing thresholds selected to favor maximum quality, minimum cost, or a balance between them. We provision sufficient replicas to serve each model's offered load without queuing; all replicas are included in the reported cost.

\noindent\textbf{Murakkab.}
Murakkab~\citep{Murakkab} uses profile-guided optimization to select workflow parameters, models, deployment configurations, and resource allocations. In our evaluated configuration, we supply the same default workflow used by the other fixed-workflow configurations and execute Murakkab's selected models and configurations using TensorRT-LLM. Because its implementation is unavailable, we reconstruct the optimizer described in the paper. Our comparison therefore characterizes this evaluated configuration and does not imply that Murakkab is inherently restricted to fixed workflows.

\noindent\textbf{{\FMNoReuse}.}
{\FMNoReuse} uses the same workflow-variant selector as {\system}, including selection across workflows, models, and deployment configurations, but deploys separate model instances for each task. Comparing {\system} with {\FMNoReuse} isolates the effect of cross-task reuse.

\noindent\textbf{Brute force and Aquatope-BO.}
For the selector study, brute-force search enumerates the complete candidate space using cached profiles and returns the minimum-cost feasible variant under Equation~\ref{eq:cost}. Aquatope-BO adapts the Bayesian-optimization strategy used for serverless workflow tuning~\citep{AQUATOPE} to our workflow-variant space. We implement it using Optuna's TPE sampler~\citep{optuna_tpe_sampler}, a budget of 200 sequential trials, and penalties for SLO violations. This is an adaptation of Aquatope's search strategy rather than an execution of the complete Aquatope system.

\noindent\textbf{Default workflows.}
For configurations receiving a supplied workflow, we use the simplest candidate for each task, composed of stages that each perform one operation (e.g., translate $\rightarrow$ summarize). LangChain additionally receives fixed model assignments and TP${}=1$, PP${}=1$.

\noindent\textbf{Latency SLOs.}
For each task, we measure an isolated reference latency using its simplest workflow, largest compatible models, and no model parallelism. We construct four tiers: $2\times$ the reference latency (\textsc{Strict}), $3\times$ (\textsc{Moderate}), $4\times$ (\textsc{Lenient}), and $5\times$ (\textsc{Relaxed}). Unless otherwise stated, T2 and T7 use \textsc{Strict} SLOs, T3 uses \textsc{Moderate}, T1 and T5 use \textsc{Lenient}, and T4 and T6 use \textsc{Relaxed} SLOs.

\noindent\textbf{Proxy-accuracy SLOs.}
Let $b_1(m),\ldots,b_6(m)$ denote model $m$'s scores on the six Open LLM Leaderboard benchmarks. We define
\begin{equation}
q(m)=\frac{1}{6}\sum_{j=1}^{6}b_j(m),
\qquad
q(\mathcal{WV})=\min_{m\in\mathcal{WV}}q(m)
\end{equation}
Following Murakkab~\citep{Murakkab}, the \textsc{Best}, \textsc{Good}, and \textsc{Basic} SLOs are the 95th, 80th, and 50th percentiles of the achievable workflow-variant proxy scores for each task. Unless otherwise stated, T2 and T7 use \textsc{Basic}; T1, T3, and T5 use \textsc{Good}; and T4 and T6 use \textsc{Best}.

We do not independently evaluate the semantic quality of generated outputs. A request instead inherits its selected workflow variant's proxy score and satisfies the proxy-accuracy SLO when $q(\mathcal{WV})\geq\alpha$.

\noindent\textbf{Testbed.}
Experiments run on one server with eight NVIDIA RTX A6000 GPUs, each with 48GB of memory. GPU pairs are connected through NVLink, while communication between pairs uses PCIe. The server has a 64-core AMD EPYC 7763 CPU and runs Ubuntu 22.04, NVIDIA driver 575.51.03, and CUDA 12.9. All LLM inference uses TensorRT-LLM with its optimized kernels, in-flight batching, and tensor- and pipeline-parallel execution.

\noindent\textbf{Serving cost.}
We measure cluster-wide GPU-memory time and convert it to dollars:
\begin{equation}
\mathrm{Cost}
=
r_{\mathrm{GB\text{-}s}}
\int_0^T \sum_j M_j(t)\, dt
\end{equation}

where $M_j(t)$ is the GPU memory occupied by model instance $j$ at time $t$, and $r_{\mathrm{GB\text{-}s}}$ is derived from the hardware rental price~\citep{aws_lambda_pricing, MaverIQ, Serveless_Cost}. This trace-level operational metric differs from the incremental cost proxy used by the selector in Equation~\ref{eq:cost}.

\noindent\textbf{Goodput.}
For an experiment lasting $T$ minutes, cluster goodput is

\begin{equation}
\mathrm{Goodput}
=
\frac{
    \left|
    \left\{
        req :
        L_{req} \leq \lambda_{req}
        \;\wedge\;
        q(\mathcal{WV}_{req}) \geq \alpha_{req}
    \right\}
    \right|
}{T}.
\end{equation}
where $L_{req}$ is request ${req}$'s end-to-end latency and $\mathcal{WV}_{req}$ is the workflow variant used to serve it. We report goodput in requests per minute.

\section{Additional Empirical Results}
\label{app:extra_eval}

\subsection{Detailed comparison of {\system}'s performance against baselines}
\label{app:e2e_detailed}

Figure~\ref{fig:eval_e2e} compares the systems under the same seven-task workload and task-specific latency and proxy-accuracy SLOs. We evaluate both bursty code arrivals and sustained conversation arrivals across four offered-load levels.

\noindent\textbf{{\system} vs. Murakkab.}
Murakkab is our closest end-to-end baseline: it selects models and deployment configurations for a user-specified workflow and makes reuse decisions offline. At lower loads, queues remain shallow, so both systems achieve similar goodput. Nevertheless, {\system} improves goodput by up to $1.46\times$ ($1.21\times$ on average) and reduces cost by $1.84$--$3.15\times$ ($2.23\times$ on average) by considering alternative workflows. For example, for the analysis-and-reporting task, Murakkab uses a three-stage workflow, whereas {\system} selects a two-stage workflow with lower latency and cost.
The difference grows at higher loads. {\system} achieves $1.45$--$10.14\times$ Murakkab's goodput ($4.73\times$ on average) while reducing cost by $1.56$--$2.77\times$ ($1.95\times$ on average). In these experiments, Murakkab's workflow and reuse choices increase its average latency by $4.65\times$. For example, when the seventh task arrives, {\system} reuses a Qwen2.5-3B instance configured with TP${}=2$, PP${}=1$, whose existing load is 16 requests/minute against a capacity of 34. Murakkab instead reuses a Qwen2.5-14B instance configured with TP${}=1$, PP${}=1$ that is already at capacity, causing queues to build and reducing goodput.

\noindent\textbf{{\system} vs. RouteLLM.}
RouteLLM fixes each task's workflow and pre-deploys a strong/weak model pair; its routing threshold therefore exposes a quality--cost tradeoff. Its maximum-quality setting routes nearly all requests through Qwen2.5-14B, such that all requests satisfy their proxy-accuracy SLOs. However, it reserves 206--235GB of GPU memory across traces and loads, incurring $2.06$--$3.84\times$ {\system}'s cost ($2.95\times$ on average). Its goodput differs from {\system}'s by at most 7\%, except at the highest code-trace load. At that point, 14\% of {\system}'s requests violate their latency SLOs, and RouteLLM achieves $1.37\times$ higher goodput (17.02 vs.\ 12.41 RPM) at $2.06\times$ the cost.
The minimum-cost setting predominantly routes requests through Qwen2.5-7B. Its cost ranges from $0.87$--$1.23\times$ that of {\system}, but only 34.9--36.8\% of requests are assigned variants satisfying their proxy-accuracy SLOs, compared with 100\% for {\system}. Consequently, {\system} achieves $1.97$--$2.82\times$ higher goodput ($2.62\times$ on average). RouteLLM's balanced setting lies between these extremes: 62.6--65.6\% of requests satisfy their proxy-accuracy SLOs, while {\system} achieves $1.31$--$1.58\times$ higher goodput ($1.52\times$ on average). The balanced setting also incurs $1.88$--$2.94\times$ {\system}'s cost ($2.35\times$ on average).

RouteLLM attains lower request latency in all three settings because we provision sufficient replicas to eliminate queuing. This replication accounts for the higher cost of its quality-oriented settings.

\noindent\textbf{{\system} vs. LangChain.}
Across traces and loads, {\system} achieves $1.13$--$12.84\times$ LangChain's goodput ($5.19\times$ on average) and reduces cost by $2.22$--$3.99\times$ ($2.93\times$ on average). LangChain fixes workflows, models, and configurations and does not reuse model instances, so it cannot adapt its resource footprint or parallelism configuration to the cluster state.
For example, LangChain deploys 15 instances consuming 290GB to serve six tasks. The remaining task requires three 29GB Qwen-2.5-14B instances, but memory fragmentation leaves no GPU with sufficient free memory. {\system} instead serves all seven tasks using three to five model instances by selecting alternative workflows and configurations and reusing compatible deployments. Static placement also increases contention: under the code trace at 50\% load, LangChain colocates four model instances on one GPU, producing an average latency of 426s, compared with 38s for {\system}.

\vspace{-2mm}
\subsection{Scaling Under Fixed Cluster Capacity}
\label{sec:eval_scalability}
\vspace{-2mm}

\begin{figure}[h]
\centering
\includegraphics[width=0.65\linewidth]{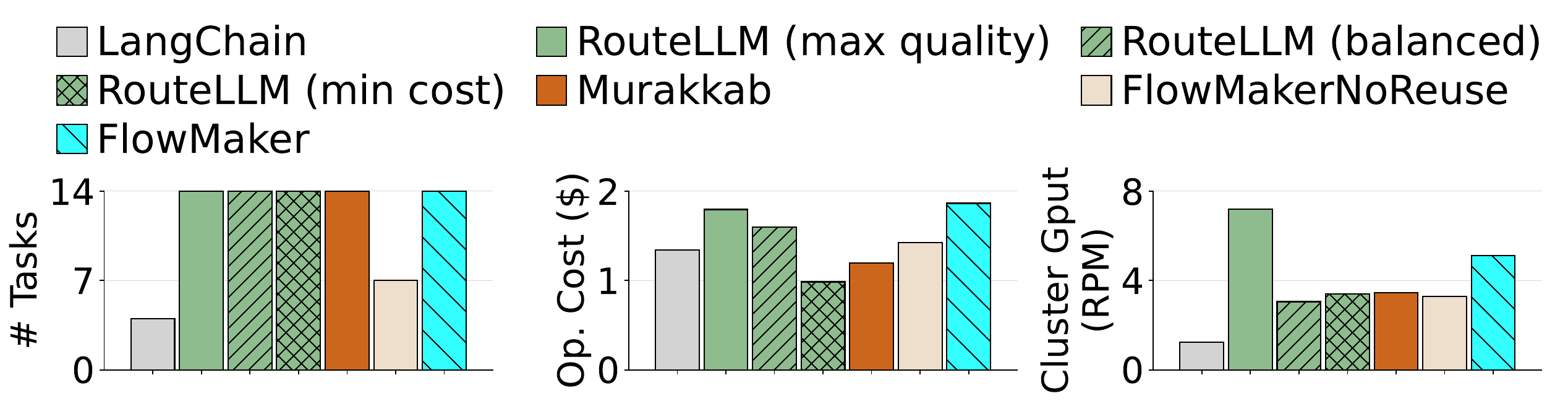}
\vspace{-3mm}
\caption{Tasks admitted (left), serving cost (middle), and SLO-compliant goodput (right) as 14 tasks arrive under fixed cluster capacity. {\system} and Murakkab admit all tasks through reuse, with {\system} achieving $1.5\times$ Murakkab's goodput. RouteLLM attains higher goodput in its maximum-quality setting through replication.
}

\label{fig:eval_scale}
\end{figure}

We next evaluate how the systems scale as tasks are added without increasing cluster capacity. We create 14 tasks by duplicating the seven-task workload and use the Azure conversation trace at 25\% load, with one task deployed every two minutes. All models are deployed within 4$\times$NVIDIA RTX A6000 GPUs. Figure~\ref{fig:eval_scale} reports the number of tasks admitted, serving cost, and aggregate SLO-compliant goodput.

LangChain and {\FMNoReuse} exhaust GPU memory after admitting four and seven tasks, respectively, because they deploy separate instances per task. {\system} and Murakkab admit all 14 tasks through reuse. However, {\system} achieves $1.5\times$ Murakkab's goodput by jointly considering alternative workflows and adapting reuse to the current load. Murakkab concentrates traffic from the 14 tasks on two shared model instances, causing contention as tasks arrive. {\system} distributes the traffic across four shared instances, preserving more serving capacity.

RouteLLM also admits all 14 tasks, since we grant it an additional advantage: it pre-deploys replicas sized to the offered load it will receive. With this headroom, its maximum-quality setting achieves $1.40\times$ {\system}'s goodput at comparable cost ($0.97\times$), as replication absorbs the queuing that {\system} incurs while deploying variants on demand. Its remaining settings trade that advantage away: the balanced and minimum-cost settings reduce cost by $1.16$--$1.90\times$, but serve only $39\%$ and $38\%$ of the requests within their proxy-accuracy SLOs, leaving goodput $1.51$--$1.68\times$ below {\system}'s. Fixed replication can therefore achieve higher goodput when the load is known ahead of time, whereas {\system} meets both latency and proxy-accuracy SLOs while deciding deployments online.

\subsection{Sensitivity of {\system} to Latency SLOs}
\label{sec:slo_sensitivity}
\vspace{-3mm}

\begin{figure}[h]
    \centering
    \includegraphics[width=0.65\linewidth]{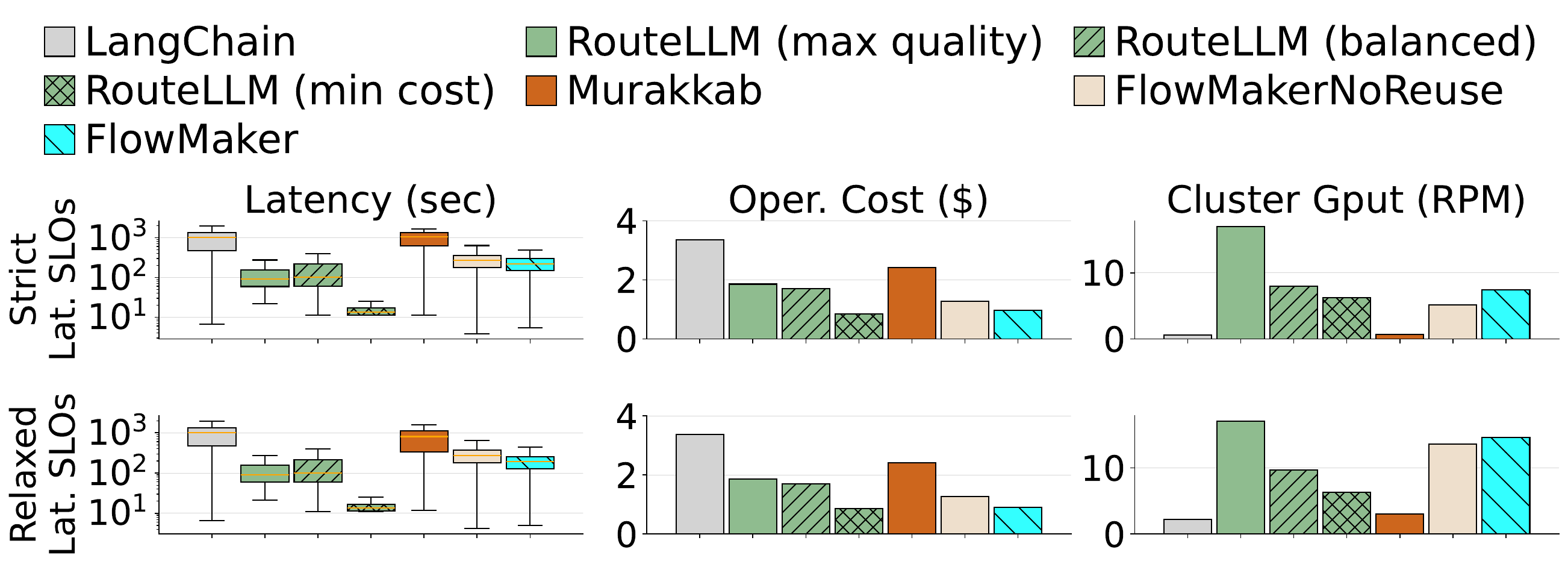}
    \vspace{-3mm}
    \caption{Per-request latency (left), serving cost (middle), and SLO-compliant goodput (right) under strict (top) and relaxed (bottom) latency SLOs, on the code trace at 100\% load. {\system} remains the most cost-efficient in both settings.
    }
    \vspace{-3mm}
    \label{fig:e2e_slo_sensitivity}
\end{figure}

In our end-to-end analysis, we evaluate {\system} under heterogeneous latency SLOs (loose, moderate, lenient, and strict). Here, we consider extreme settings where all tasks have either loose or strict SLOs, using the code trace at 100\% load with the same accuracy SLOs as in the end-to-end experiment.
Figure~\ref{fig:e2e_slo_sensitivity} shows that
all systems, except RouteLLM, experience a substantial drop in goodput under strict SLOs—approximately halving relative to loose SLOs—due to the difficulty of meeting tight constraints under limited resources. However, {\system} remains the most robust, improving goodput by $1.1$–$12.4\times$ and reducing cost by $1.3$–$3.8\times$ across those baselines in both settings.
RouteLLM is the exception: replication keeps its latency low, so its goodput is identical under both SLO settings. 
Under strict SLOs, this exceeds {\system} by $2.29\times$ and $1.08\times$ for the max-quality and balanced variants, at $1.94\times$ and $1.77\times$ the cost, while the min-cost variant reduces cost by $1.13\times$ and goodput by $1.18\times$ compared to {\system}.

\subsection{{\system} Incurs Minimal Overheads}
\label{sec:overheads}
Section~\ref{sec:eval_algo-efficiency} shows our search algorithm efficiently selects workflow-variants; here, we quantify model loading overheads. Loading takes 1–5s (depending on size), amounting to only 0.4–2.5\% of SLO time. These costs are small relative to {\system}’s gains from on-the-fly reuse (1.13$\times$ goodput, 1.42$\times$ cost reduction), making the tradeoff worthwhile. Moreover, these overheads can be further reduced using model-loading techniques from prior work~\citep{DynamoLLM, ServerlessLLM, InferX}.

\end{document}